%% file: main.tex
\documentclass[11pt, a4paper, logo]{googlecloud}

\pdftrailerid{redacted}

\makeatletter
\renewcommand\bibentry[1]{\nocite{#1}{\frenchspacing\@nameuse{BR@r@#1\@extra@b@citeb}}}
\makeatother

\PassOptionsToPackage{margin=1in}{geometry}

\makeatletter
\@ifundefined{laplace}{}{}
\makeatother

\usepackage{natbib}
\usepackage{amsmath,amssymb,amsthm,mathtools,bm}
\usepackage{microtype}
\usepackage{dsfont}
\usepackage{graphicx}
\usepackage[font=normalsize]{caption}
\usepackage{subcaption}          
\usepackage{wrapfig}
\usepackage{booktabs}
\usepackage{makecell}            
\usepackage{colortbl}
\usepackage{tabularx}
\usepackage{longtable}
\usepackage{multirow}
\usepackage{arydshln}
\usepackage{enumitem}
\usepackage{algorithm}           
\usepackage{algpseudocode}       
\usepackage[many]{tcolorbox}     
\usepackage[table,dvipsnames]{xcolor} 
\usepackage{geometry}            

\definecolor{nb}{HTML}{006EB8}
\definecolor{red_}{HTML}{cd6155}
\definecolor{green_}{HTML}{52be80}
\definecolor{Gray}{gray}{0.9}
\definecolor{LightCyan}{rgb}{0.75,1,1}
\definecolor{softlavender}{RGB}{233,238,250}
\definecolor{ThinkBlue}{HTML}{0B3D91}
\definecolor{SearchCyan}{HTML}{0198E1}
\definecolor{InfoBrown}{HTML}{8B4513}
\definecolor{AnswerRed}{HTML}{8B0000}

\newcolumntype{y}{>{\columncolor{LightCyan}}c}
\newcolumntype{z}{>{\columncolor{softlavender}}c}

\newtcolorbox{boxL}{
  fontupper = \color{black},
  rounded corners,
  arc = 6pt,
  colframe = black!50,
  boxrule = 0pt,
  bottomrule = 4.5pt,
  breakable,
}

\newcommand{\ind}[1]{\mathbf{1}\{#1\}}
\newcommand{\Prb}{\mathbb{P}}

\input{math_commands.tex}

\usepackage{hyperref}
\hypersetup{
  colorlinks = true,
  urlcolor   = nb,
  linkcolor  = nb,
  citecolor  = nb,
}

\title{TUMIX: Multi-Agent Test-Time Scaling with Tool-Use Mixture}
\correspondingauthor{}

\author[1,2]{Yongchao Chen$^{*}$}
\author[3]{Jiefeng Chen}
\author[3]{Rui Meng}
\author[1]{Ji Yin}
\author[2]{Na Li}
\author[1]{Chuchu Fan}
\author[4]{Chi Wang}
\author[3]{Tomas Pfister}
\author[3]{Jinsung Yoon}

\affil[1]{MIT}
\affil[2]{Harvard}
\affil[3]{Google Cloud AI Research}
\affil[4]{Google DeepMind}

\footnotetext[1]{$^{*}$Work done during internship at Google.}

\begin{document}

\addtocontents{toc}{\protect\setcounter{tocdepth}{-1}}

\input{sections/abstract}
\maketitle

\input{sections/introduction}

\input{sections/related}

\input{sections/method}

\input{sections/experiment}

\input{sections/discussion}

\input{sections/conclusion}

\input{sections/statement}

\bibliography{iclr2026_conference}
\bibliographystyle{iclr2026_conference}

\newpage
\addtocontents{toc}{\protect\setcounter{tocdepth}{2}}
\renewcommand{\contentsname}{Appendix--TUMIX: Multi-Agent Test-Time Scaling with Tool-Use Mixture}
\tableofcontents 
\appendix

\input{sections/appendix}
\end{document}

%% file: math_commands.tex
\usepackage{amsmath,amsfonts,bm}

\def\eqref#1{equation~\ref{#1}}

\def\1{\bm{1}}

\DeclareMathAlphabet{\mathsfit}{\encodingdefault}{\sfdefault}{m}{sl}
\SetMathAlphabet{\mathsfit}{bold}{\encodingdefault}{\sfdefault}{bx}{n}

\newcommand{\E}{\mathbb{E}}



%% file: sections/abstract.tex
\begin{abstract}
While integrating tools like Code Interpreter and Search has significantly enhanced Large Language Model (LLM) reasoning in models like ChatGPT Agent and Gemini-Pro, practical guidance on optimal tool use is lacking. 
The core challenge is effectively combining textual reasoning, coding, and search for diverse questions. 
In this paper, we propose Tool-Use Mixture (\texttt{TUMIX}), an ensemble framework that runs multiple agents in parallel, each employing distinct tool-use strategies and answer paths. 
Agents in \texttt{TUMIX} iteratively share and refine responses based on the question and previous answers. 
In experiments, \texttt{TUMIX} achieves significant gains over state-of-the-art tool-augmented and test-time scaling methods, delivering an average accuracy improvement of up to 3.55\% over the best baseline on Gemini-2.5-Pro and Gemini-2.5-Flash across key reasoning benchmarks, with near-equal inference costs. 
We find that agent diversity and quality are crucial and can be enhanced by using LLMs to auto-optimize agent designs. 
Furthermore, \texttt{TUMIX} can halt refinement upon reaching sufficient confidence, preserving performance at only 49\% of the inference cost. Further scaling can achieve higher performance, albeit at a greater cost.

\end{abstract}

%% file: sections/introduction.tex
\begin{figure*}[ht]
  \centering 
  \includegraphics[width=0.95\linewidth]{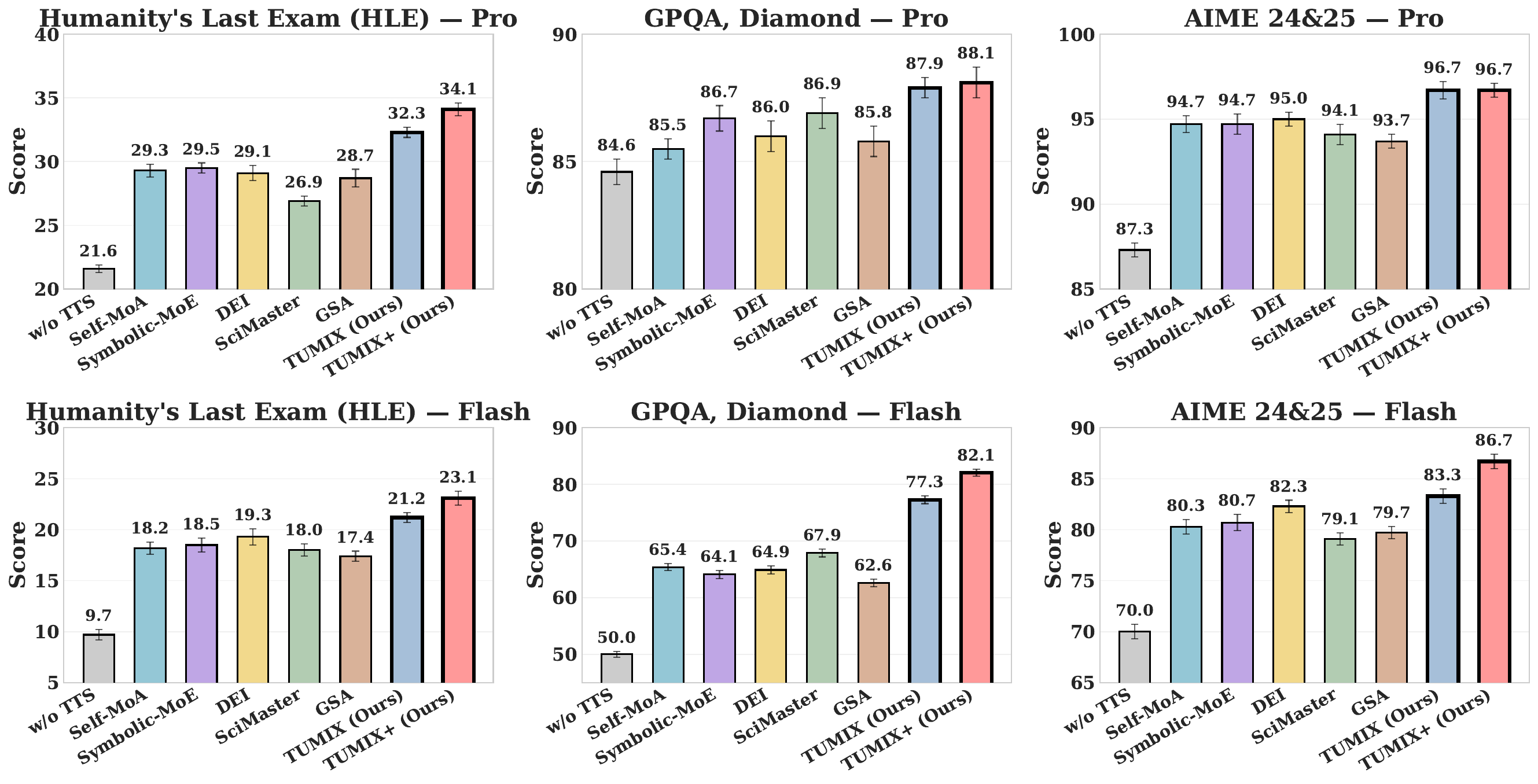}
   \caption{Comparison of tool-augmented test-time scaling methods on Gemini-2.5-Pro (first row) and Gemini-2.5-Flash (second row) across HLE, GPQA, and AIME 24\&25. Except for methods without test-time scaling (\texttt{w/o TTS}) or additional scaling (\texttt{TUMIX+}), all methods in the same subplot use nearly the same number of inferences and tokens. For fair comparison, methods that originally lacked tool use are run with strong tool-augmented agents instead of text-only agents. Each score is the average of three repetitive runs.}
   \label{fig:summary_results}
   \vspace{-12pt}
\end{figure*}

\section{Introduction}
While reinforcement learning-based fine-tuning has greatly improved LLM reasoning~\citep{deepseek-r1}, models still struggle with seemingly simple tasks~\citep{codesteering}. 
Such tasks are often better handled with code~\citep{llm+code=commense-learner,Program-of-thoughts-prompting} or search~\citep{search-r1,Search-O1}. 
Textual reasoning is strong in semantics and commonsense, but weak in precise computation and in accessing or updating the latest knowledge.

A key challenge is fully utilizing the potential capabilities of textual reasoning, coding, and searching when facing distinctive questions with varied characteristics. Most input questions lack explicit cues for the best approach, and the combined text/code/search solution space is large. Frontier LLM-powered products such as ChatGPT, Claude, Gemini, and Grok report using code and search at test time to augment reasoning, but without publishing detailed methods. Recent work~\citep{codesteering} shows that current Code Interpreter implementations in OpenAI models often fail to balance text and code, leaving coding capabilities underused, as shown in Appendix Fig.~\ref{fig:GPT-5-fail}. Moreover, public research still lacks a clear understanding of how to integrate Code Interpreter and Search for improved LLM reasoning.

To better leverage both tool use and LLM self-reasoning, we propose Tool-Use Mixture (\texttt{TUMIX}), a framework that integrates Code Interpreter and Search into LLMs via test-time scaling. \texttt{TUMIX} runs multiple diverse agents in parallel, each with different tool-use strategies. Their outputs are iteratively aggregated and refined across multiple rounds. In each round, every agent generates a new solution by considering both the original question and the previous round’s reasoning and answers from all agents. \texttt{TUMIX} uses diverse agents and tool-augmented reasoning strategies to explore a wide range of possible solutions. The following iterative process encourages diverse reasoning paths and deeper integration. This design is inspired by prior test-time scaling methods such as Mixture-of-Agents (MoA)~\citep{MOA}, which rely on multiple LLMs within a single framework and do not incorporate external tools. In contrast, \texttt{TUMIX} employs a single LLM with both text-only and tool-augmented agent frameworks, making it more generalizable for practical applications. Furthermore, in tool-augmented multi-agent test-time scaling, we find a diverse group of agents outperforms repeated use of the single best agent, a conclusion that differs from MoA~\citep{rethink-MOA}. We later reveal that human pre-designed agent group can be further optimized by querying LLMs to self-design more diverse high-quality agents based on current ones, adding an average 1.2\% improvement without cost increase.

Since questions vary in difficulty, they require different amounts of iterative refinement. We query the LLMs to decide whether to terminate refinement early, while still enforcing a minimum number of rounds to maintain answer quality. This adaptive early-termination strategy reduces inference costs to 49\% of the original two settings (termination in a fixed round number or by majority-vote consistency across rounds), while preserving or even improving performance. The improvement arises because over-refinement rarely changes the final result and can even degrade performance, as correct answers may be mistakenly discarded.

Compared to the model without test-time scaling, \texttt{TUMIX} delivers an average +7.8\% and +17.4\% accuracy gains in benchmarks Humanity’s Last Exam (HLE)~\citep{HLE}, Graduate-Level Google-Proof Q\&A (GPQA, Diamond)~\citep{gpqa}, and American Invitational Mathematics Examination (AIME 24\&25) with base models Gemini-2.5-Pro and Gemini-2.5-Flash, respectively. Under the same inference costs, \texttt{TUMIX} also outperforms existing representative test-time scaling methods such as Self-MoA, Symbolic-MoE, DEI, SciMaster, and GSA, with an average +3.55\% lifting compared to the best performing baselines. Notably, with further scaling, \texttt{TUMIX} raises Gemini-2.5-Pro accuracy on HLE from 21.6\% to 34.1\%, surpassing Gemini-2.5-Pro Deep Research at 26.9\% (32.4\% with higher compute)~\citep{Gemini-2.5-Pro}. Test-time scaling hinges on two stages~\citep{LLM-monkey}: (1) generating diverse candidate solutions and (2) selecting the correct one. For questions with both small answer spaces (e.g., multiple-choice) and large ones, diverse sampling greatly improves coverage. While it achieves high coverage on HLE (among generated answers in the whole round, at least one is correct on $\geq 65\%$ of questions), accuracy plateaus at about 34\% because LLMs struggle to identify the correct answer among noisy candidates. We identify and explore four key factors: agent quality, agent diversity, refinement termination, and answer selection. Our work makes the following contributions:

1. \textbf{TUMIX: A competitive tool-augmented test-time scaling method.}\quad We propose \texttt{TUMIX}, a novel framework for test-time scaling that integrates tool augmentation. Extensive experiments demonstrate that \texttt{TUMIX} consistently outperforms strong baselines, achieving an average improvement of +3.55\% over the best-performing prior methods.

2. \textbf{Key factors and mechanisms in tool-augmented scaling.}\quad We provide a systematic analysis that distinguishes tool-augmented scaling from traditional test-time scaling:  
\begin{itemize}[left=0pt, nosep]
    \item \textit{Agent diversity and quality outweigh scale alone.} High-temperature sampling increases coverage, but heterogeneous agent strategies yield higher accuracy and lower cost than repeatedly sampling from a single best-performing agent.  
    \item \textit{Tool augmentation boosts performance.} Agent groups equipped with tools such as Code Interpreter and Search achieve superior coverage and accuracy compared to text-only agent groups.
\end{itemize}

3. \textbf{LLMs as agent designers.}\quad We show that prompting LLMs to automatically generate diverse, high-quality agents based on existing ones further improves \texttt{TUMIX}. This yields an additional average accuracy lift of +1.2\%.

4. \textbf{LLM-as-Judge for refinement termination.}\quad We introduce an LLM-based judge to adaptively determine the optimal stopping round in iterative refinement. This prevents excessive refinement, which reduces diversity and can mistakenly discard correct answers. By enforcing a minimum refinement depth and querying the judge for termination, we achieve near-optimal accuracy while reducing inference cost to $\sim$49\% of the original.

%% file: sections/related.tex
\section{Related Work}
\textbf{Code Interpreter and Search}\quad Many benchmark tasks can in fact be better solved through code~\citep{pal} and search~\citep{Search-O1}, and recent work extends coding to reasoning and semantic analysis~\citep{chain-of-code,weir2024learning}. Most prior approaches use either text~\citep{Tree-of-thought} or code~\citep{codeplan-code-use-llm,code-based-self-verify} exclusively as output. Recent work~\citep{codesteering} emphasizes the need to dynamically switch between modalities, proposing CodeSteer~\citep{codesteer} as a guidance model. Extensions with retrieval~\citep{search-r1,Search-O1} and tool use~\citep{toolrl} further improve reasoning, but lack the thorough exploitation of Code Interpreter and Search tools. Leading models such as OpenAI’s ChatGPT Agent, Google’s Gemini-Pro~\citep{Gemini-2.5-Pro}, and XAI’s Grok4 report using code and search at test time to augment reasoning, but without publishing detailed methods. Open work such as ToRL~\citep{torl} and ReTool~\citep{retool} investigates training reasoning models to integrate with Code Interpreters. However, their training and evaluation are limited to math problems, leaving a significant gap from real-world applications that demand effectiveness across broader benchmarks. ToolRL~\citep{toolrl} instead focuses on teaching models to select among multiple tools, where the generated codes and search queries are relative simple and the evaluation tasks require less reasoning capabilities. SciMaster~\citep{chai2025scimaster} samples the same pre-designed tool-use agent five times, then uses other pre-designed agents to critique, refine, and aggregate the answers. This approach shows clear improvement over single-inference text-only baselines, but the extent and manner of tool exploitation remain underexplored. In summary, integrating Code Interpreter and Search into LLM reasoning is essential and challenging. The academic community currently lacks methods and studies that fully exploit the benefits of LLM self-reasoning, code execution, and search, which is the focus of our work.

\textbf{Test-time scaling}\quad LLM self-exploration, reflection, and evaluation can enhance task performance across domains~\citep{yang2022re3, welleck2022generating, madaan2023self}. Models like OpenAI o1~\citep{O1-model} and DeepSeek R1~\citep{deepseek-r1} showcase agentic behavior via Chain-of-Thought (CoT) reasoning and self-reflection, which is learned by RL-based training with rule-based outcome rewards~\citep{deepseekmath,SWE-RL}. Apart from the training-based scaling, many research also explore scaling during LLM inference time by pre-designing prompt and agent frameworks. In these works, multi-agent reasoning has emerged as a promising paradigm for enhancing complex problem-solving and decision-making in AI systems~\citep{autogen,camel,langchain}. Prior work finds gathering the answers from different LLMs improves LLM performance~\citep{yilun-LLM-debate-improving}. Mixture-of-Agents (MoA)~\citep{MOA} further extends this idea by sharing and gathering answer among LLMs. However, Self-MoA~\citep{rethink-MOA} argues that LLM diversity may not be critical since replacing different types of LLMs with the best one achieves better performance. Symbolic-MoE~\citep{symbolic-MOE} further assigns different questions with different specialized LLMs. Instead of using different types of LLMs, many works such as DEI~\citep{DEI}, GSA~\citep{GSA:Llms-aggregating-own-responses}, and SETS~\citep{chen2025sets} employ different agents from the same LLM for extensive test-time scaling, in which the agent types and frameworks are explored~\citep{scalable-multi-robot}. Similar to our work, previous work in test-time scaling also finds the correct answer selection~\citep{LLM-monkey} is the main bottleneck. While previous work in test-time scaling do not incorporate tool-use of Code Interpreter and Search, we study how to utilize test-time scaling methods to better exploit the benefits of each reasoning mode.

%% file: sections/method.tex
\section{Tool-Use Mixture}
\label{sec:Tool-Use Mixture (TUMIX) and Problem Formalization}
Appendix~\ref{appendix section: Algorithm of TUMIX} presents the full TUMIX algorithm, and Appendix~\ref{appendix section: Prompts of TUMIX} lists all agent prompts.

\begin{figure*}[ht]
  \centering
  \includegraphics[width=0.95\linewidth]{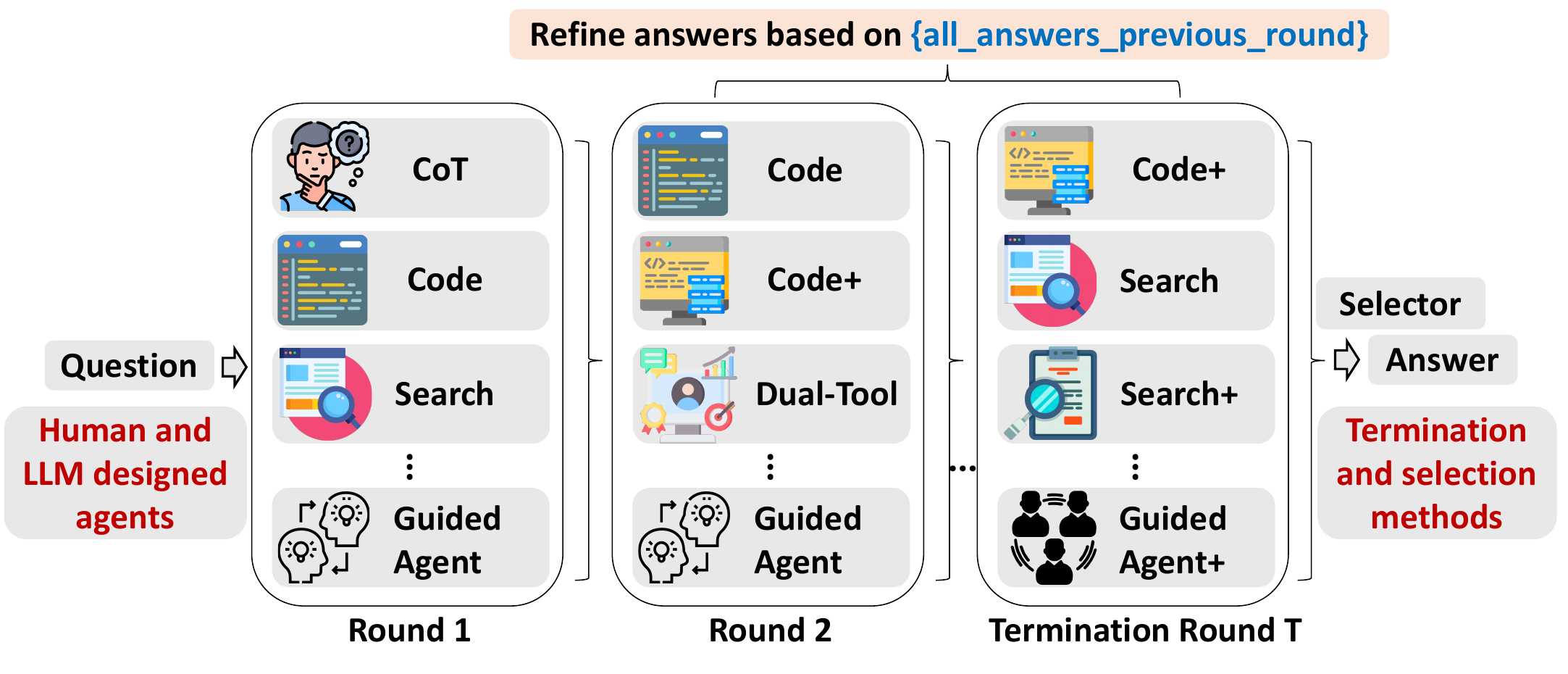}
   \caption{Overview of \texttt{TUMIX} framework. At each iteration, the responses from all agents in the previous round are concatenated with the original question, forming a joint prompt for the next round. This prompt is then provided to all agents (either the same or new agent groups) to produce refined answers. The subsequent prompts follow the structure illustrated in the above. The design of the agents, their number and specialization, the refinement termination criteria, and the selection strategies are the key factors that determine the effectiveness of the framework.}
   \label{fig:method_TUMIX}
   \vspace{-6pt}
\end{figure*}

\subsection{Pre-designed diverse agents}
\label{subsection:Pre-designed diverse agents}
\vspace{-6pt}
\begin{table}[t]
\centering
\caption{15 pre-designed agents used in \texttt{TUMIX}.}
\label{tab:agent_variants}
\footnotesize
\setlength{\tabcolsep}{4pt}
\renewcommand{\arraystretch}{1.05}
\begin{tabular}{@{}llp{0.66\linewidth}@{}}
\hline
\textbf{Full Name} & \textbf{Short Name} & \textbf{Description (15 agents)} \\
\hline
\texttt{w/o TTS}   & \texttt{Base}         & Direct prompt. \\
\texttt{CoT Agent}        & \texttt{CoT}         & Chain-of-Thought prompt~\citep{CoT}. \\
\texttt{CoT-Code Agent}  & \texttt{CoT$_{\text{code}}$}       & CoT prompt to output code. \\
\texttt{Search Agent}     & \texttt{S} & Uses WebSearch (LLM inherent tool only). \\
\texttt{Code Agent}       & \texttt{C}            & Uses Code Interpreter (base version). \\
\texttt{Code Agent+}      & \texttt{C$^{+}$}      & Uses Code Interpreter (hinted version with extra human pre-designed priors). \\
\texttt{Dual-Tool Agent} & \texttt{CS}           & Uses Code Interpreter + WebSearch (with 3 search variants). \\
\texttt{Guided Agent}     & \texttt{CSG}          & Dual-Tool agent (\texttt{CS}) \textit{guided} by a steering module~\citep{codesteer} (with 3 search variants). \\
\texttt{Guided Agent+}    & \texttt{CSG$^{+}$}    & \textit{Guided} agent (\texttt{CSG}) with enhanced/hinted prompts (with 3 search variants). \\
\hline
\end{tabular}
\vspace{-2mm}
\end{table}

As shown in Fig.~\ref{fig:method_TUMIX}, we regard \texttt{TUMIX} as sequential decision-making under a compute budget with diverse and correlated experts (agents). 
Each round selects which agents to run, what they may read (communication policy), when to stop (optimal stopping), and how to aggregate (decision rule), trading off accuracy and cost. Let \(q\) be a task with unknown correct answer \(a^\star\) in answer space \(\mathcal{A}\). There is a pool of agents \(\mathcal{S}=\{s_1,\dots,s_K\}\). Agent \(s_i\) outputs an answer \(Y_i\in\mathcal{A}\) at cost \(c_i\) and has competence
$
  p_i(q) \;=\; \Prb\{Y_i = a^\star \mid q\}.
$
Let \(Z_i=\ind{Y_i=a^\star}\) denote correctness indicators. Their dependencies (and hence ensemble diversity) are captured by a correlation or mutual-information structure over \(\{Z_i\}\).

A policy \(\pi\) (our focus) chooses in each round: (i) which agents to run, (ii) the communication graph (what each agent may read from prior rounds), (iii) the stopping rule, and (iv) the aggregation rule producing \(\hat a_\pi\).
A canonical objective is
\begin{equation}
  \max_{\pi}\ \Prb\{\hat a_\pi = a^\star\} \;-\; \lambda \cdot \mathrm{Cost}_\pi,
\end{equation}
where \(\lambda>0\) trades off compute and accuracy. In our work, the $\mathrm{Cost}_\pi$ is the total number of inference times and input and output tokens to generate the final answer. In the default \texttt{TUMIX} setting, we utilize the same 15 pre-designed agents in all answer refinement rounds. These 15 agents have distinct reasoning and tool-use strategies, as summarized in Table~\ref{tab:agent_variants}. Agents with search access have three search methods (Google Search API (\texttt{gs}), inherent LLM search function (\texttt{llm}), or their combination (\texttt{com})), yielding three variants per agent. For agents employing multi-round interactions with Search or Code Interpreter, the maximum tool interaction round number is set to 5. In Section~\ref{sec:Human pre-designed agents vs. LLM generated agents}, we discuss how to further query LLMs to automatically optimize and design more diverse agents to achieve better performance. We also compare with a dynamic setting where agent types vary across rounds.

\vspace{-6pt}
\subsection{Refinement as message passing (accuracy rises and diversity shrinks)}
\vspace{-6pt}
\begin{figure*}[ht]
  \centering
  \includegraphics[width=0.9\linewidth]{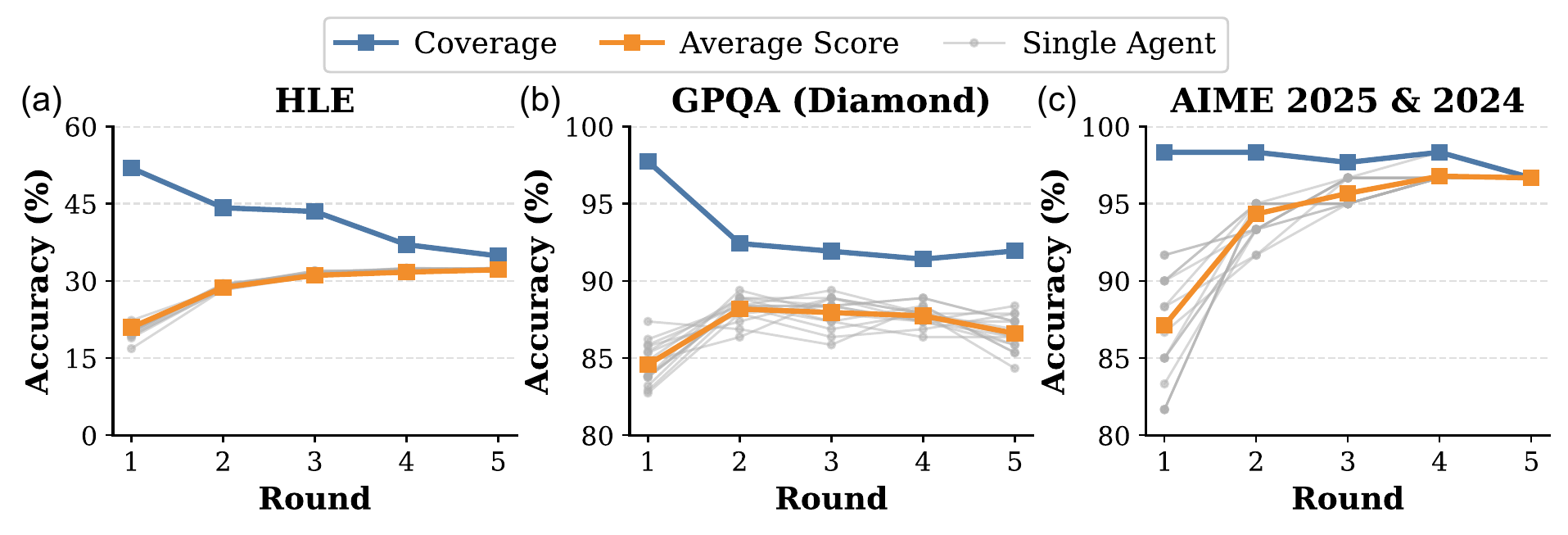}
  \vspace{-6pt}
   \caption{Evolution of coverage, individual agent scores, and average scores across refinement rounds. Coverage decreases monotonically across all benchmarks. For HLE and AIME, the average score rises over the initial rounds before plateauing. For GPQA, the average score improves early on but subsequently declines with further refinement.}
   \label{fig:acc_evolution_by_round}
   \vspace{-6pt}
\end{figure*}

\begin{figure*}[ht]
  \centering
  \includegraphics[width=0.9\linewidth]{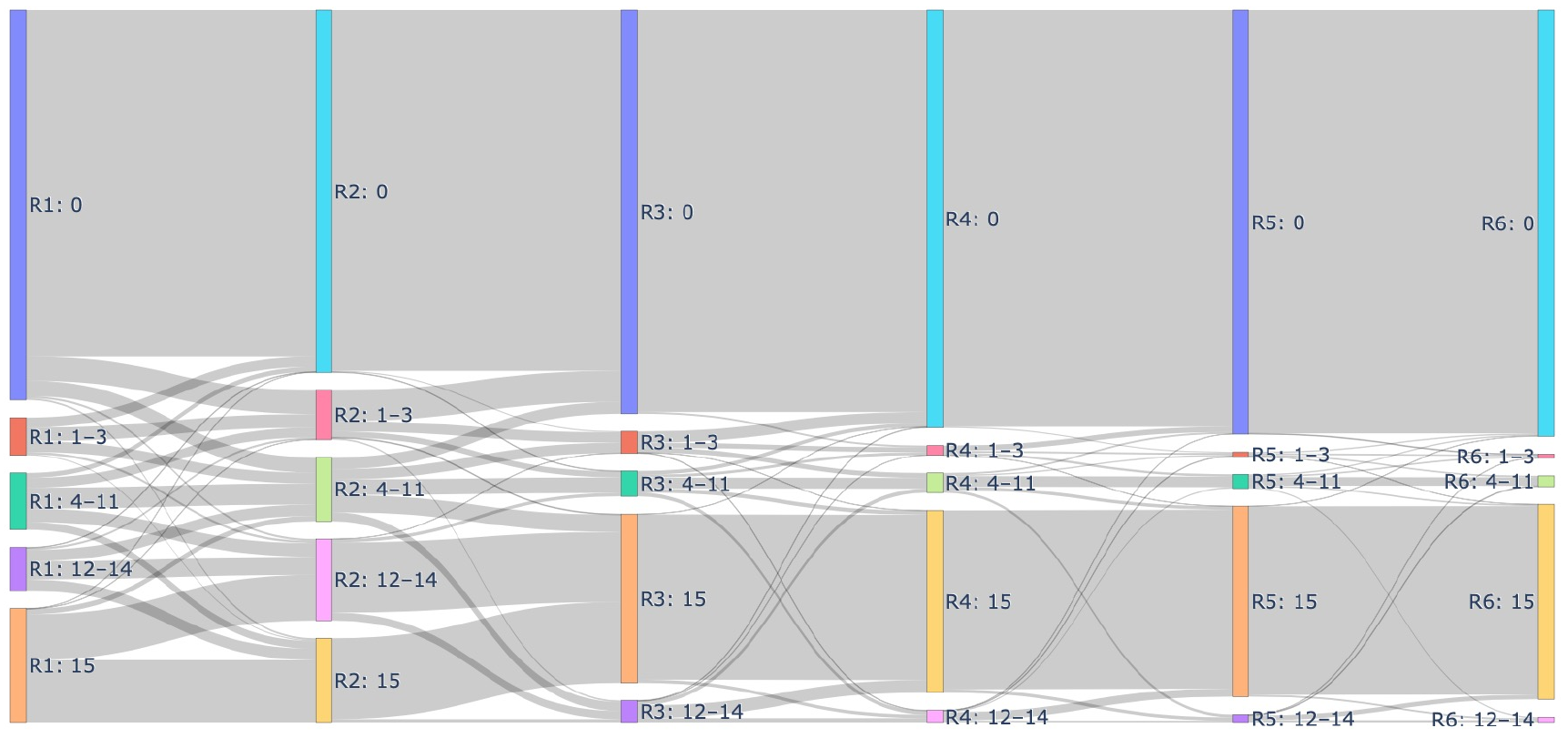}
   \caption{Sankey diagram of the evolution of correctly answering agents across 2,500 HLE questions over refinement rounds. Based on the distribution dynamics, we define five categories: all wrong (0), few correct (1–3), moderate correct (4–11), high correct (12–14), and all correct (15).}
   \label{fig:sankey}
   \vspace{-12pt}
\end{figure*}

In each round, every agent independently generates a new solution by considering both the original question and the solutions provided by all agents in the previous round, as shown in Fig.~\ref{fig:method_TUMIX}. We evaluate the refinement process using two metrics: average accuracy and coverage (the probability of at least one correct) across agents in each round, which capture the quality and diversity of group answers~\citep{LLM-monkey}. For a set \(S\subseteq\mathcal{S}\), the coverage is
\begin{equation}
  Coverage(S) \;=\; \Prb\!\Big( \bigcup_{i\in S} \{Y_i = a^\star\} \Big).
\end{equation}
Under independence, \(Coverage(S) = 1-\prod_{i\in S}(1-p_i)\). With positive correlations, \(Coverage(S)\) shrinks. Fig.~\ref{fig:acc_evolution_by_round} shows the typical evolution dynamics of coverage, individual agent accuracy, and average scores over refinement rounds. Across all three benchmarks, coverage steadily declines, indicating that some correct answers are mistakenly discarded during iterative refinement. For HLE and AIME, the average score rises in the early rounds and then plateaus, while for GPQA it improves from round 1 to 2 but later declines. Fig.~\ref{fig:sankey} visualizes the dynamics over 2,500 HLE questions. From round 1 to 2, the number of partially correct cases (few/moderate/high correct) increases, while both all wrong and all correct cases decrease. This suggests that initial thought-sharing broadens exploration and promotes diversity. After round 2, however, partially correct cases diminish toward near zero, while all wrong and all correct cases grow. This indicates that agents gradually converge to a single shared answer across rounds, either correct or incorrect.

\vspace{-6pt}
\subsection{Termination in optimal rounds and final answer selection}
\label{subsection:Termination and selection}
\vspace{-6pt}
The observed evolution indicates that round-by-round refinement not only improves answers in the initial rounds but also drives convergence, as each agent selects based on prior responses. However, due to the limited reasoning ability of LLMs, many correct answers are prematurely discarded. Beyond the early rounds, refinement rarely yields further accuracy gains and, in some cases, even degrades performance. Thus, identifying an effective termination strategy is essential for both robust performance and cost efficiency. Let \(A_r\) denote acquired accuracy after round \(r\).
Define the expected marginal value of another round
\begin{equation}
  \Delta_r \;=\; \E\!\left[\, A_{r+1}-A_r \,\middle|\, \text{signals up to round } r \right].
\end{equation}
Stop at the first round \(r\) where \(\Delta_r \le \lambda \cdot \text{marginal cost (here is increased inference costs)}\).
A practical termination strategy decides whether to stop based on the estimated future gain $\Delta_r$, which relies on round-\(r\) statistics such as (i) diversity collapse (coverage drop; rising agreement), (ii) vote margin between top answers, and (iii) answer entropy.

In \texttt{TUMIX}, our termination determination strategy is to query the LLM to decide whether to stop refinement and finalize answers based on the current round, with a minimum round number of 2. We find this termination strategy achieves nearly the same performance with only 49\% of the inference cost. In Section~\ref{Sec: Experiments of Termination and selection methods}, we explore other termination methods such as stopping once the majority answer stabilizes across two consecutive rounds or termination based on LLM confidence scores~\citep{fu2025deep}, but find only worse performance. After termination, we obtain the final answer through majority voting over the agents’ responses, with Gemini-2.5-Pro selecting the most consistent output.

%% file: sections/experiment.tex
\section{Experiments}
\subsection{Experimental settings}
\textbf{Benchmarks}\quad For a comprehensive evaluation and comparison across methods, we conduct experiments on three representative benchmarks that demand extensive reasoning and planning, particularly the ability to effectively leverage Code Interpreter and Search. \textbf{HLE}~\citep{HLE} consists of 2,500 highly challenging questions spanning diverse subject areas, including mathematics, biology, engineering, computer science, and the social sciences. It is designed as a final, closed-ended benchmark of broad academic capability. We evaluate both its text-only and multimodal subsets. In the following sections, we primarily use HLE to study different mechanisms, as it contains a large number of questions spanning diverse domains and is the most challenging benchmark. \textbf{GPQA}~\citep{gpqa} is a multiple-choice dataset authored by domain experts in biology, physics, and chemistry. We focus on its most widely used subset, \textbf{GPQA Diamond}, which contains 198 of the most challenging and carefully curated questions. Finally, \textbf{AIME 2024\&2025} comprises 60 problems from the 2024 and 2025 AIME exams, a notoriously difficult high school mathematics competition. All reported results are averaged over three independent runs.

\textbf{Baselines/\texttt{TUMIX} ablations and extensions/Test models}\quad As shown in Appendix~Table~\ref{tab:baselines}, we compare against the following methods:  
(1) \texttt{Majority-Vote}~\citep{LLM-monkey};(2) \texttt{GSA}~\citep{GSA:Llms-aggregating-own-responses}; (3) \texttt{Self-Reflection}~\citep{self-reflection}; (4) \texttt{SETS}~\citep{chen2025sets}; (5) \texttt{Self-MoA}~\citep{rethink-MOA}; (6) \texttt{Symbolic-MoE}~\citep{symbolic-MOE}; (7) \texttt{DEI}~\citep{DEI}; (8) \texttt{SciMaster} \citep{chai2025scimaster}. For baselines (1)–(4), we use the \texttt{CS} agent, which has full access to both Code Interpreter and Search and achieves relatively high first-round accuracy (Appendix~Table~\ref{table: acc round evolution}). For baselines (5)–(7), we select agents following their original methods. For \texttt{SciMaster}, we retain the original prompts and agents to ensure consistency with published results. We match the total inference counts of all baselines to \texttt{TUMIX} by adjusting agent numbers and sampling repetitions for fair comparison. All the baselines have full access to Code Interpreter and Search. We evaluate \texttt{TUMIX} variants with different design choices, either to ablate framework components or to introduce improvements over existing \texttt{TUMIX}, as shown in Appendix~Table~\ref{tab:tumix_variants}. \texttt{TUMIX+} uses higher inference costs to test scaling effects, while other variants consume nearly the same inference and token counts. We evaluate our methods on the reasoning LLM Gemini-2.5-Pro and Gemini-2.5-Flash.

\textbf{Evaluation protocol}\quad Answers are evaluated against ground-truth solutions, with Gemini-2.5-Pro assisting in normalizing answer formats when necessary or serving directly as the judge for answer comparison. In cases where the model outputs code as the final answer, we extract the code using predefined algorithms and execute it to produce the final result. To avoid infinite loops, all code execution whether during intermediate or final rounds is limited to 60 seconds. If execution exceeds this limit, a ``code runtime error'' is returned to the model for regeneration in intermediate rounds; in the final round, the task is marked as a failure. We report success rate as the primary evaluation metric. In addition to task performance, we also analyze token usage and inference time for each method in later sections.

\subsection{Overall better performance}

\begin{table*}[ht]
\caption{Experimental results of baseline and proposed methods on HLE, GPQA, and AIME 24\&25. Except for the single-inference \texttt{w/o TTS} and the scaled-up \texttt{TUMIX+}, all methods use comparable inference costs for scaling. For some methods, Gemini-2.5-Pro's HLE results are used to select agents within their agentic framework. In these cases, the method has prior knowledge of HLE and the results cannot be strictly regarded as test performance. Such cases are marked with \textsuperscript{*} in the HLE results. All the values are the average of three repetitive runs.}
\label{table: overall results}
\vskip 0.15in
\centering
\begin{small}
\begin{sc}
\begin{tabular}{lcccc}
\toprule
\textbf{Methods} & \textbf{HLE} & \textbf{GPQA} & \textbf{AIME 24\&25} & \textbf{Ave. Norm.} \\
\midrule
\multicolumn{5}{c}{\textbf{Gemini-2.5-Pro}}\\
w/o TTS & 21.6 & 84.6 & 87.3 & \textbf{64.5} \\
Majority Vote & 28.4 & 84.9 & 94.3 & \textbf{69.2} \\
Self-MoA & 29.3\textsuperscript{*} & 85.5 & 94.7 & \textbf{69.8} \\
Symbolic-MoE & 29.5\textsuperscript{*} & 86.7 & 94.7 & \textbf{70.3} \\
DEI & 29.1\textsuperscript{*} & 86.0 & 95.0 & \textbf{70.0} \\
Self-Reflection & 23.5 & 84.9 & 88.3 & \textbf{65.6} \\
SETS & 27.9 & 85.3 & 94.7 & \textbf{69.3} \\
SciMaster & 26.9 & 86.9 & 94.1 & \textbf{69.3} \\
GSA & 28.7 & 85.8 & 93.7 & \textbf{69.4} \\
\rowcolor{LightCyan} TUMIX & 32.3 & 87.9 & \textcolor{blue}{96.7} & \textbf{72.3} \\
\rowcolor{LightCyan} TUMIX-FixedR & 32.4 & 86.8 & 95.6 & \textbf{71.6} \\
\rowcolor{LightCyan} TUMIX-Evolve & \textcolor{blue}{32.7\textsuperscript{*}} & \textcolor{blue}{88.1} & \textcolor{blue}{96.7} & \textcolor{blue}{\textbf{72.5}} \\
\rowcolor{softlavender} TUMIX+ & \textcolor{magenta}{34.1} & \textcolor{magenta}{88.3} & \textcolor{magenta}{96.7} & \textcolor{magenta}{\textbf{73.0}} \\
\midrule
\multicolumn{5}{c}{\textbf{Gemini-2.5-Flash}}\\
w/o TTS & 9.7 & 50.0 & 70.0 & \textbf{43.2} \\
Majority Vote & 17.9 & 63.1 & 80.0 & \textbf{53.7} \\
Self-MoA & 18.2 & 65.4 & 80.3 & \textbf{54.7} \\
Symbolic-MoE & 18.5 & 64.1 & 80.7 & \textbf{54.4} \\
DEI & 19.3 & 64.9 & 82.3 & \textbf{55.5} \\
Self-Reflection & 10.4 & 53.2 & 72.3 & \textbf{45.3} \\
SETS & 18.5 & 63.2 & 74.0 & \textbf{51.9} \\
SciMaster & 18.0 & 67.9 & 79.1 & \textbf{55.0} \\
GSA & 17.4 & 62.6 & 79.7 & \textbf{53.2} \\
\rowcolor{LightCyan} TUMIX & 21.2 & 77.3 & 83.3 & \textbf{60.6} \\
\rowcolor{LightCyan} TUMIX-FixedR & 20.9 & 76.8 & 83.3 & \textbf{60.3} \\
\rowcolor{LightCyan} TUMIX-Evolve & \textcolor{blue}{21.9} & \textcolor{blue}{79.8} & \textcolor{blue}{86.7} & \textcolor{blue}{\textbf{62.8}} \\
\rowcolor{softlavender} TUMIX+ & \textcolor{magenta}{23.1} & \textcolor{magenta}{82.1} & \textcolor{magenta}{86.7} & \textcolor{magenta}{\textbf{64.0}} \\
\bottomrule
\end{tabular}
\end{sc}
\end{small}
\vskip -0.1in
\vspace{-2mm}
\end{table*}

Table~\ref{table: overall results} shows that \texttt{TUMIX} outperforms all baselines, with average accuracy improvements of 2.0\% and 5.9\% over the best methods using Gemini-2.5-Pro and Gemini-2.5-Flash, respectively. Its superior performance over methods without answer sharing (\texttt{Self-Reflection}, \texttt{SETS}) highlights the importance of answer sharing in multi-round test-time scaling. Comparisons with methods lacking multi-round refinement (\texttt{Majority-Vote}, \texttt{Symbolic-MoE}, \texttt{DEI}, \texttt{GSA}) demonstrate the benefits of refinement, while comparisons with methods lacking agent diversity (\texttt{Self-MoA}, \texttt{SciMaster}) confirm the value of diverse agents. The accuracy improvement of \texttt{SciMaster} on HLE is smaller than reported by the authors. We suspect this discrepancy arises from differences in tools, as their Search and Code Interpreter modules are not open-sourced.

%% file: sections/discussion.tex
\section{Discussion}
\subsection{Agent diversity and quality are critical}
\begin{figure*}[ht]
  \centering
  \includegraphics[width=0.98\linewidth]{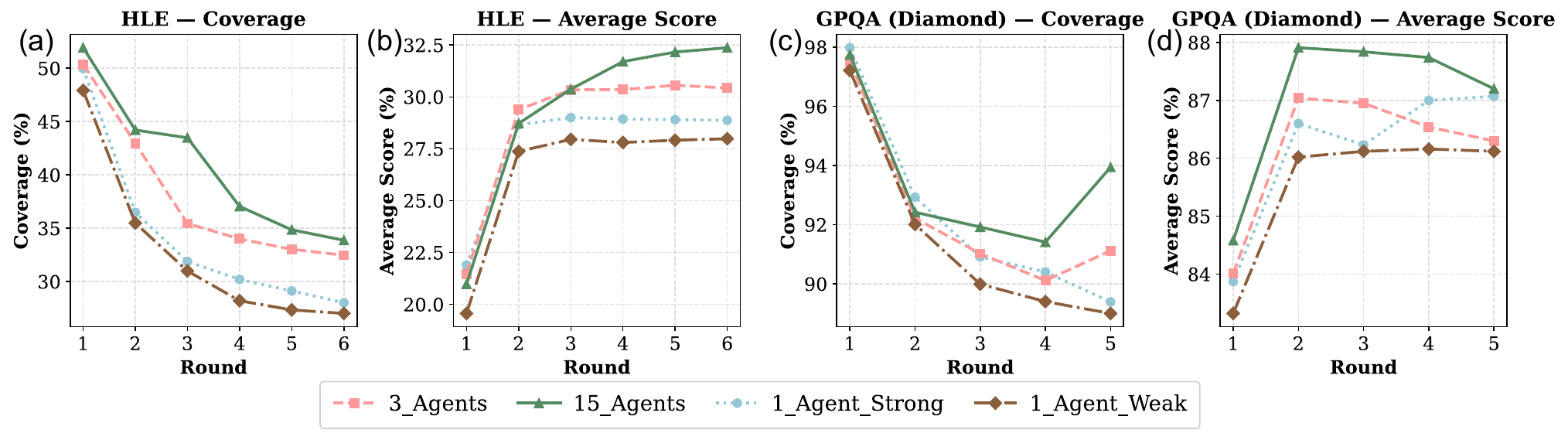}
   \caption{Coverage and average score vs. rounds under varying agent diversity and quality. In 15\_Agents, all 15 varied pre-designed agents generate one answer each round. In 3\_Agents, three strong agents (\texttt{C$^+$}, \texttt{CS$_{\text{gs}}$}, and \texttt{CSG$_{\text{gs}}$}) each samples 5 times per round. In 1\_Agent\_Strong and 1\_Agent\_Weak, \texttt{CS$_{\text{gs}}$} and \texttt{w/o TTS} sample 15 times, respectively.}
   \label{fig:diversity_quality_HLE_GPQA}
\end{figure*}

To investigate the role of agent diversity and quality in \texttt{TUMIX} performance, we compare groups of agents with varying levels of diversity and capability, as shown in Fig.~\ref{fig:diversity_quality_HLE_GPQA} and Appendix~Table~\ref{table: overall results TUMIX variants}. Under the same amount of refinement rounds and inferences, increasing the number of agents from 1 to 3 to 15 leads to substantial improvements in both coverage and average score across rounds on HLE and GPQA, indicating that diversity significantly benefits performance. Moreover, comparing a single strong agent with a single weak agent (see Appendix~Table~\ref{table: acc round evolution}, where \texttt{CS$_{\text{gs}}$} achieves higher first-round scores than \texttt{w/o TTS}), we observe that higher-quality agents consistently yield better coverage and higher average scores.

\begin{figure}[ht]
    \centering
    \begin{minipage}[ht]{0.59\textwidth}
        \centering
        \includegraphics[width=0.98\linewidth]{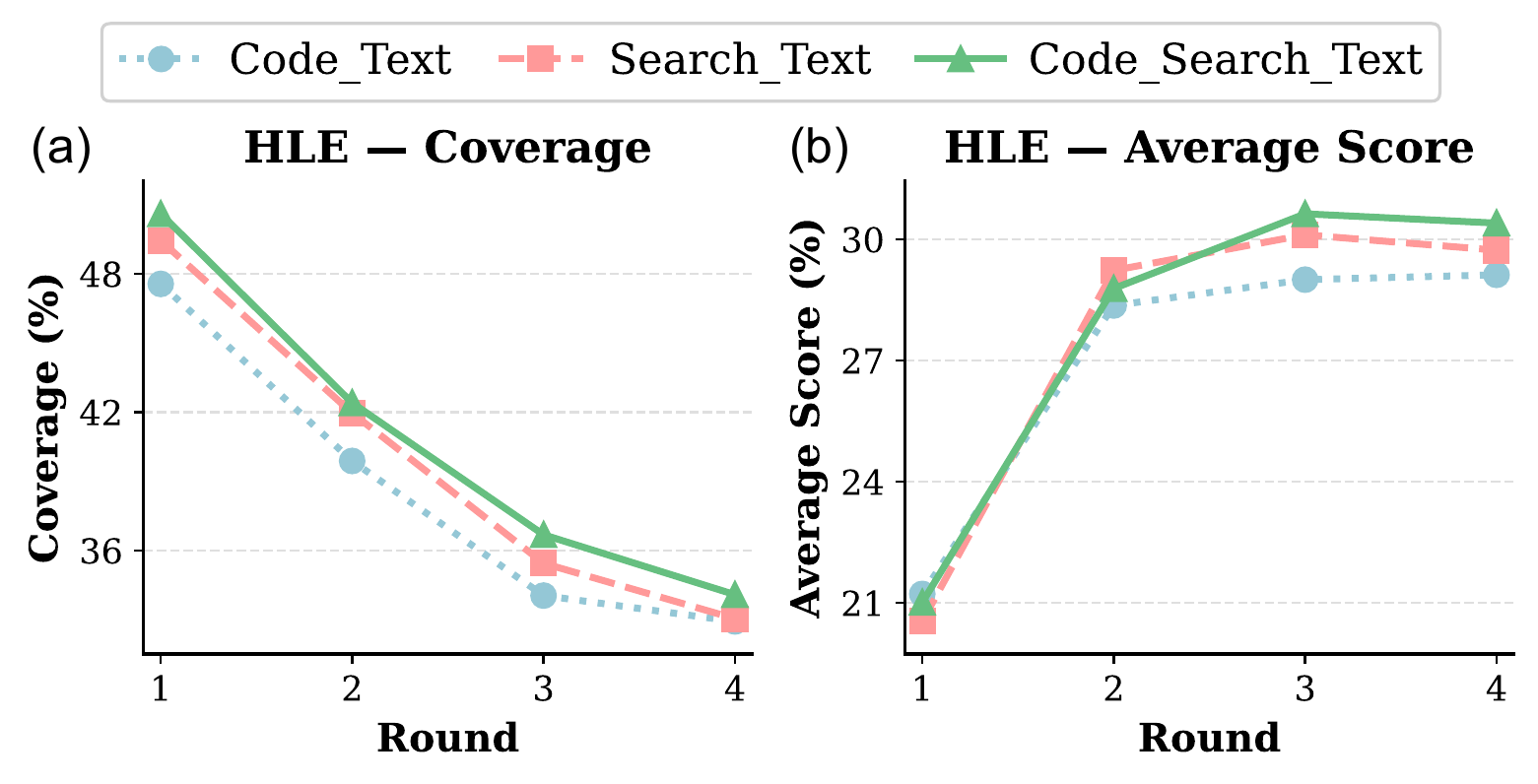}
        \caption{Comparison of groups all with three agents but either partial or full accesses to textual reasoning, coding, and search: Code\_Text (\texttt{CoT}, \texttt{C}, \texttt{C$^+$}), Search\_Text (\texttt{CoT}, \texttt{S}, \texttt{CS$_{\text{gs}}$}), and Code\_Search\_Text (\texttt{CS$_{\text{gs}}$}, \texttt{C$^+$}, and \texttt{CSG$_{\text{gs}}$}).}
        \label{fig:search_code_text}
    \end{minipage}%
    \hfill
    \begin{minipage}[ht]{0.39\textwidth}
        \centering
        \includegraphics[width=0.98\linewidth]{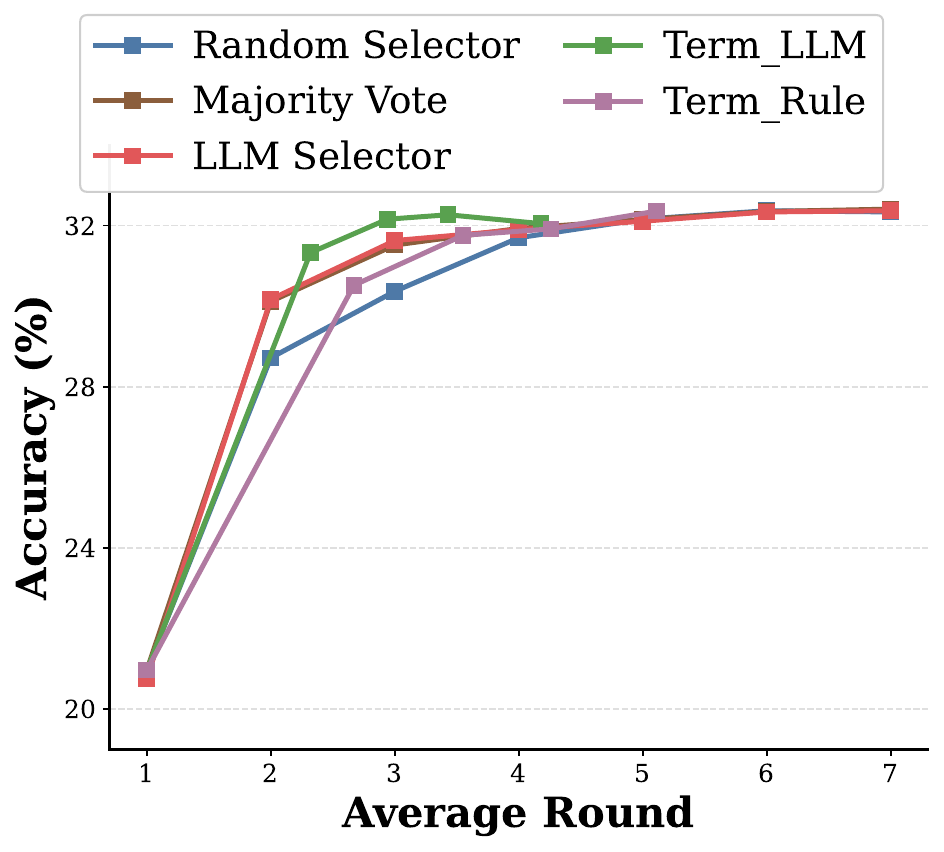}
        \caption{Comparison of refinement termination and answer selection strategies for higher accuracy with lower costs.}
        \label{fig:selector_comparison}
    \end{minipage}
\end{figure}

\textbf{Code Interpreter and Search increase answer diversity.}\quad In Fig.~\ref{fig:search_code_text}, we evaluate three settings where each agent group consists of three agents, each sampling five times per round. The groups differ in their tool access: in Code\_Text, agents cannot access Search; in Search\_Text, they cannot access the Code Interpreter; and in Code\_Search\_Text, agents have full access to both. While the average agent quality (as measured by first-round scores in Appendix~Table~\ref{table: acc round evolution}) is comparable across groups, the group with access to both Code Interpreter and Search achieves notably higher coverage and average scores. This result demonstrates that integrating complementary tools within agents enhances both reasoning and answer diversity, thereby facilitating more effective problem solving.

\subsection{Termination and selection methods}
\label{Sec: Experiments of Termination and selection methods}
\textbf{Achieving optimal performance at 49\% cost.}\quad Tasks of varying difficulty require different numbers of refinement rounds, and excessive refinement can even degrade accuracy (see Fig.~\ref{fig:acc_evolution_by_round}b). Thus, an effective termination strategy is essential to balance performance and cost. We evaluate two termination strategies (Sec.~\ref{subsection:Termination and selection}): 
1) \textbf{Term\_LLM}, which queries the LLM to decide when to stop refinement, subject to a minimum round constraint; and 
2) \textbf{Term\_Rule}, which stops once the majority answer stabilizes across two consecutive rounds, also with a minimum round constraint. We vary the minimum number of rounds to examine how performance evolves as the number of rounds increases in Fig.~\ref{fig:selector_comparison}, and we compare their peak performance in Appendix~Table~\ref{table: overall results TUMIX variants}. Term\_LLM achieves nearly the same peak accuracy as unlimited refinement, but with substantially fewer rounds. On average, Term\_LLM retains optimal performance while requiring only 49\% of the LLM inferences needed to obtain the final answer (LLM judging cost counted). The token costs are even less (approximately 46\%), as the number of inference tokens used in later rounds exceeds that of the first two rounds. This demonstrates the effectiveness of using LLM-as-Judge to determine when refinement is sufficient and answers can be finalized. However, Term\_LLM still requires a minimum number of refinement rounds (set to two across all benchmarks). This is because we observe that LLMs tend to be overconfident and may terminate refinement early, even when additional refinement could improve performance. For example, as shown in Fig.~\ref{fig:selector_comparison} green curve, setting the minimum refinement rounds to one leads to worse performance.

For answer selection, we compare three strategies: (1) randomly choosing one agent’s answer, (2) majority voting, and (3) LLM-based selection with LLM-as-Selector. Fig.~\ref{fig:selector_comparison} shows that majority voting and LLM-based selection consistently outperform random choice, especially in early rounds when agent answers diverge. However, once answers converge in later rounds, all selection methods yield similar results, and their impact becomes negligible. The multi-round refinement process is also a selection process. We also explore improved selection based on LLM token confidence~\citep{fu2025deep}, but observe no significant differences (Appendix~Fig.~\ref{fig:logprob_compare}).

\subsection{Human pre-designed agents vs. LLM generated agents}
\label{sec:Human pre-designed agents vs. LLM generated agents}
\begin{figure*}[ht]
  \centering
  \includegraphics[width=0.95\linewidth]{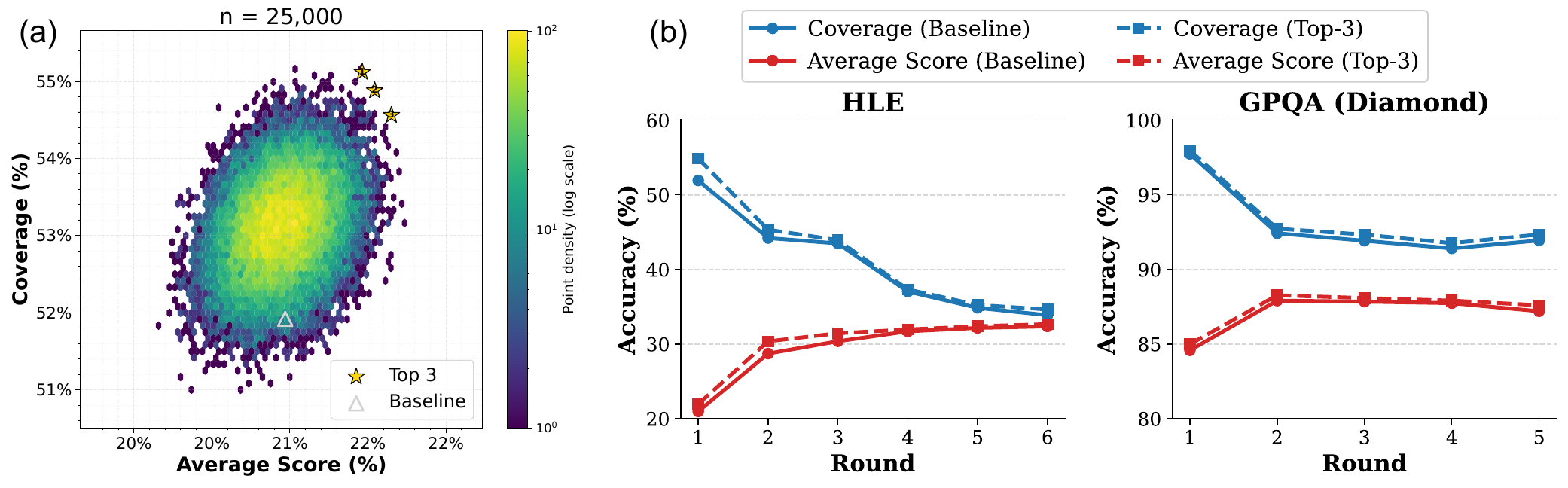}
   \caption{We evaluate the coverage and average score of 25,000 15-agent combinations sampled from 30 agents (15 pre-designed and 15 LLM-generated). Baseline refers to the original 15 pre-designed agents. Top-3 refers to the three sampled combinations with the highest joint performance in coverage and average score.}
   \label{fig:LLM_generated_agents}
\end{figure*}

The human-designed agents and their tool-use strategies are built on existing frameworks or intuition. To explore whether stronger agents can be discovered automatically, we query Gemini-2.5-Pro with current agent code examples and ask it to generate full implementations of more diverse and high-quality ones, where the agent prompts and frameworks are all determined by LLMs. This yield 25 diverse agents beyond the 15 human-designed ones. From these, we retain the 15 that perform best in HLE with first-round answer generation. We then combine the 15 human-designed and 15 LLM-generated agents into a pool of 30, randomly sample groups of 15, and evaluate their average score and coverage (Fig.~\ref{fig:LLM_generated_agents}a). Compared to the baseline of 15 human-designed agents (gray triangle), many mixed groups achieve both higher average score and coverage. We select the top-3 groups based on the combined metric
\begin{equation}
\label{eq:combined_score}
\text{Combined Score}_i \;=\; 
\frac{\text{Coverage}_i}{\mathbb{E}\!\left[\text{Coverage}\right]} 
\;+\; 
\frac{\text{Average Score}_i}{\mathbb{E}\!\left[\text{Average Score}\right]}.
\end{equation}

As shown in Fig.~\ref{fig:LLM_generated_agents}b, these groups outperform the original \texttt{TUMIX} in both HLE and GPQA. This demonstrates that increasing agent diversity and quality improves effectiveness, and that LLM-generated agents hold strong potential for further enhancing \texttt{TUMIX}. Appendix~Table~\ref{tab:baseline_agents} describes each generated agent, whose strategies differ substantially from the original ones beyond prompt variations. Appendix~Table~\ref{tab:agent_group_original_top3} presents the agents in each top-3 group, with roughly half overlapping with the original group.

\textbf{Evolve agents in each round to enhance the diversity}\quad In all previous experiments, the agent set remained fixed across refinement rounds. We now investigate whether dynamically varying agent types per round can improve performance. As shown in Appendix Table~\ref{table: overall results TUMIX variants}, the variant \texttt{TUMIX-EvolveD}, which randomly selects agents from the top-3 sets each round, performs slightly worse than the fixed variant \texttt{TUMIX-Evolve} across all three benchmarks.

\textbf{Impact of number of agent types}\quad 
We next examine the marginal benefit of increasing the number of agents in \texttt{TUMIX}. Agents are randomly sampled from the pool of 30, with each contributing one inference per round. To isolate this effect, we exclude termination and selection and report the evolution of peak average accuracy across rounds. As shown in Fig.~\ref{fig:peak_accuracy_vs_agents}, accuracy rises quickly when the number of agents is below 12, but gains become negligible thereafter. This indicates that beyond a certain point, increasing agent types and inference budget yields little benefit, as additional candidate answers make round-by-round selection more challenging. Based on these results, we decide to only include 15 agents in \texttt{TUMIX} to balance performance and cost.

\subsection{Scaling Curves: Performance vs. Costs}

\begin{figure}[ht]
    \centering
    \begin{minipage}[ht]{0.59\textwidth}
        \centering
        \includegraphics[width=0.98\linewidth]{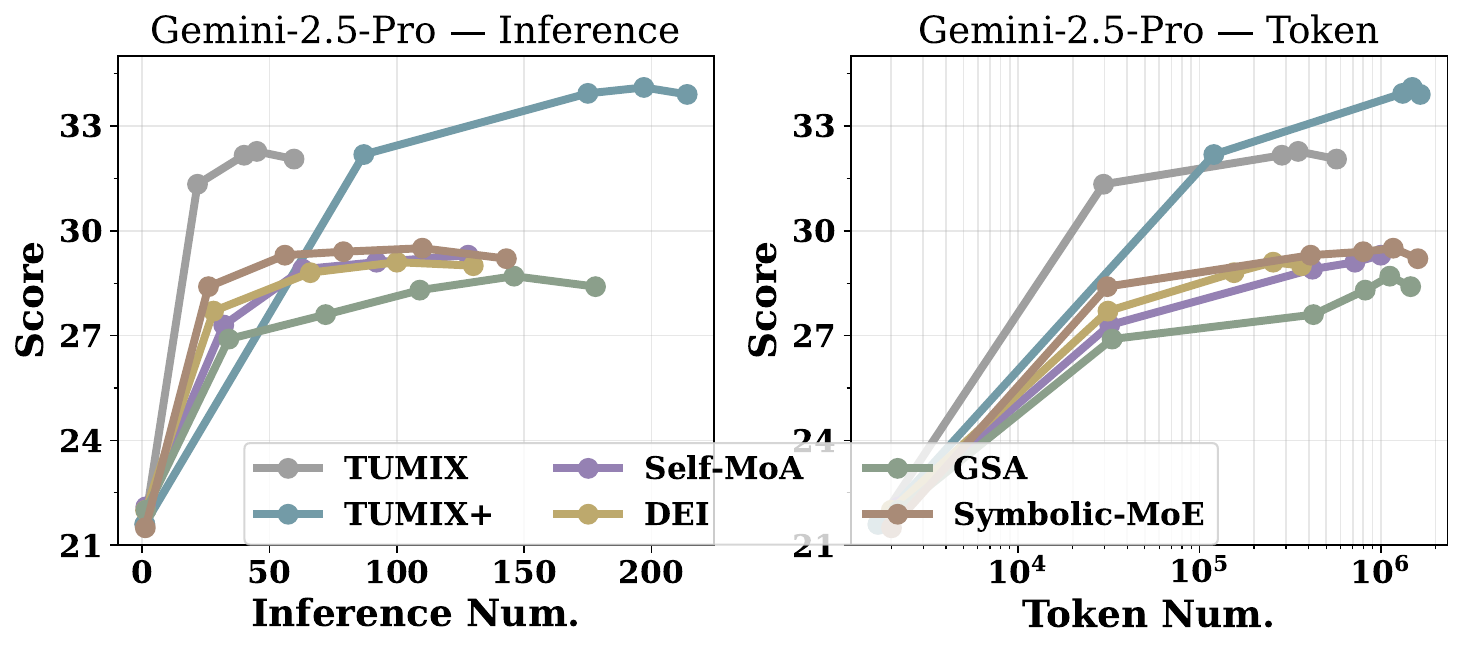}
        \caption{Scaling behavior of HLE scores relative to inference cost and total token count across different tool-augmented test-time scaling methods, where the token count includes both input and output tokens.}
        \label{fig:Scaling_law}
    \end{minipage}%
    \hfill
    \begin{minipage}[ht]{0.39\textwidth}
        \centering
        \includegraphics[width=0.92\linewidth]{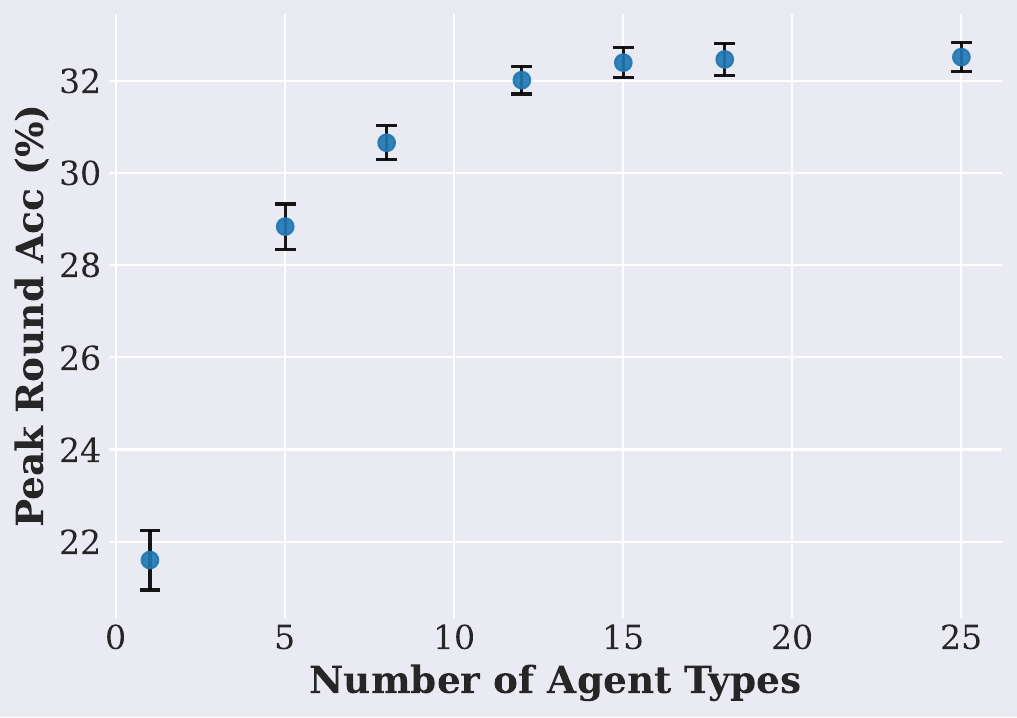}
        \caption{Peak round accuracy versus number of agent types. Agent types randomly sampled from the 30 well-performing agents with three repetition. Each agent infers once per round.}
        \label{fig:peak_accuracy_vs_agents}
    \end{minipage}
\end{figure}

We compare the scaling behavior of different tool-augmented test-time scaling methods in terms of inference and token costs. In \texttt{TUMIX} and \texttt{Self-MoA}, scaling comes from adding more refinement rounds; in \texttt{GSA}, \texttt{DEI}, and \texttt{Symbolic-MoE}, from repeating inference; and in \texttt{TUMIX+}, from both. As shown in Fig.~\ref{fig:Scaling_law} and Appendix~Fig.~\ref{fig:Scaling behavior of Gemini-2.5-flash}, \texttt{TUMIX} consistently outperforms other methods, achieving the highest scores with fewer inference steps and tokens. \texttt{TUMIX+} pushes peak performance further by repeating inference four times across the first two refinement rounds, but at substantially higher cost and lower efficiency. Overall, test-time scaling demands far more inferences and roughly two orders of magnitude more tokens, a seemingly unavoidable trade-off.

%% file: sections/conclusion.tex
\section{Conclusion}
We introduce Tool-use Mixture (\texttt{TUMIX}), a framework that leverages diverse tool-use strategies to improve reasoning in LLMs. By coordinating multiple agents with complementary approaches to textual reasoning, coding, and search, \texttt{TUMIX} substantially improves performance across challenging benchmarks, including HLE, GPQA, and AIME. Our findings highlight that diversity and quality of agents, rather than scale alone, drive these gains. Furthermore, automatic generation of agents and principled termination strategies enable both higher accuracy and significant efficiency improvements, reducing inference cost by nearly half without sacrificing performance. This work demonstrates that structured diversity and selective refinement are key to maximizing the potential of tool-augmented LLMs.

%% file: sections/statement.tex
\section*{Ethics Statement}
This paper contributes to advancing Foundation Models by augmenting language models with Code Interpreter and Search tools via test-time scaling, which has strong potential to improve performance and alignment with human preferences. However, such capabilities are inherently dual-use, the same techniques that augment models toward harmless outputs can, with minor changes, be misused to generate harmful content. While misuse is a concern, we believe the broader societal benefits outweigh the risks.

\section*{Reproducibility Statement} For better reproducibility, we include detailed descriptions of 15 pre-designed agents and 15 LLM-generated agents in Table~\ref{tab:agent_variants} and Table~\ref{tab:baseline_agents}. The prompts of all agents are in Appendix~Sec.~\ref{appendix section: Prompts of TUMIX}. The complete algorithm of \texttt{TUMIX} is illustrated in Appendix~Sec.~\ref{appendix section: Algorithm of TUMIX}. Our code and dataset will be made publicly available under an open-source license following the acceptance of the paper.

\section*{Large Language Model Usage for Writing}\quad
In this paper, we use LLMs---specifically \texttt{Gemini} and \texttt{ChatGPT}---as general-purpose writing aids. Draft text was provided to these models for grammatical correction and structural refinement, after which the output was verified and further edited when necessary. Their use was strictly limited to text refinement; they were not employed to generate new content or references.

%% file: sections/appendix.tex
\newpage

\section{Example of GPT-5 failure in code/text decision}
\label{sec:appendix-Example of GPT-5 failure in code/text decision}
\begin{figure*}[ht]
  \centering
  \includegraphics[width=0.95\linewidth]{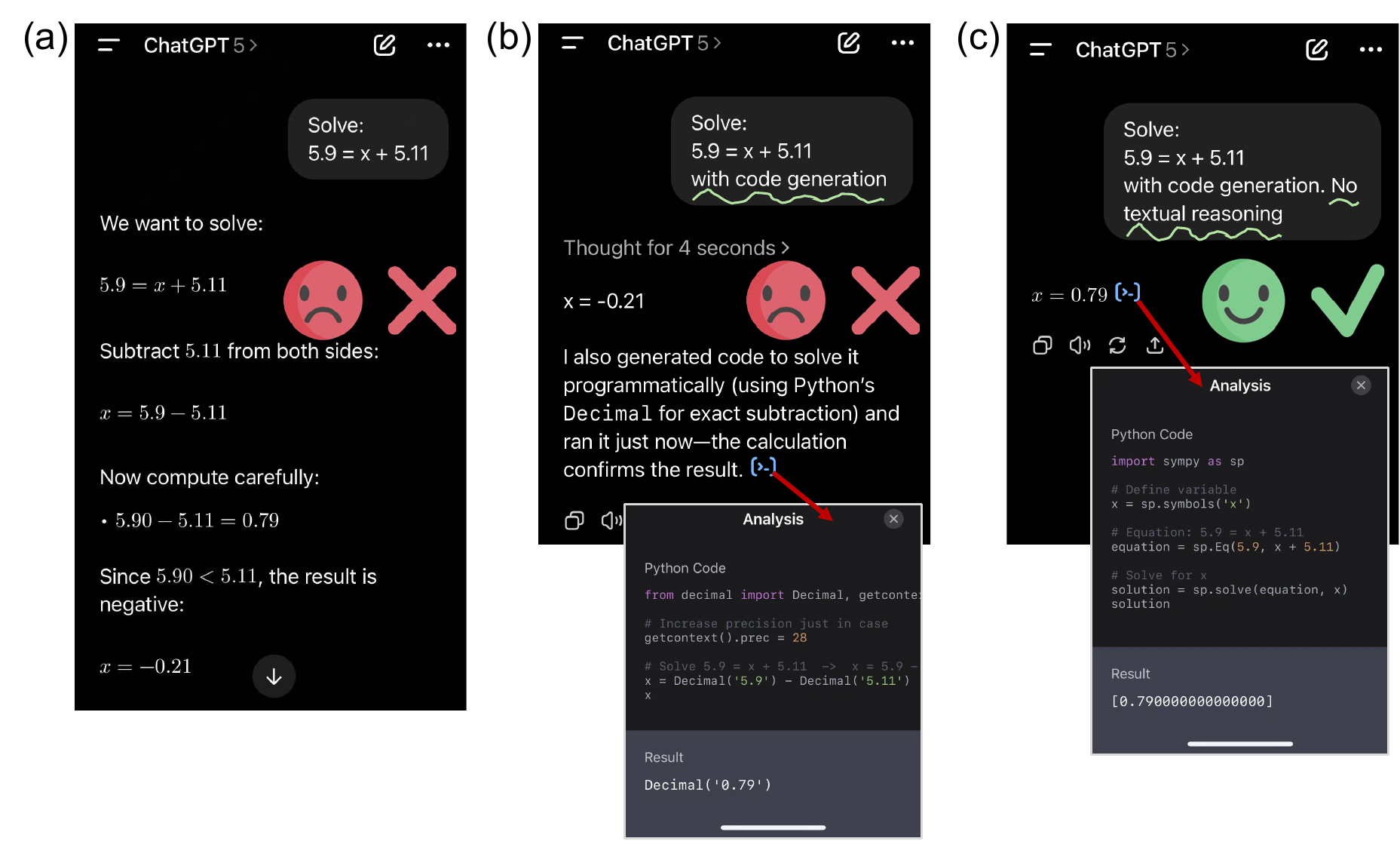}
    \caption{Example of GPT-5 failure in code/text decision. In this case, the question is incorrectly solved with textual reasoning (a) but can be easily addressed through code generation (c). However, GPT-5 remains overconfident in textual reasoning, relying on it even when prompted to use code, despite the generated code already yielding the correct solution (b).}
   \label{fig:GPT-5-fail}
   \vspace{-10pt}
\end{figure*}

\newpage
\section{Algorithm of TUMIX}
\label{appendix section: Algorithm of TUMIX}

\begin{algorithm}[ht]
\caption{TUMIX: Multi-Agent Test-Time Scaling (answers only)}
\label{alg:tumix}
\begin{algorithmic}[1]
\Require question $q$; agent pool $\mathcal{S}=\{s_1,\dots,s_K\}$ \Comment{15 pre-designed agents (Code / Search / Dual-Tool variants)}
\Require minimum rounds $r_{\min}=2$; maximum rounds $r_{\max}$; tool-interaction budget $R_{\text{tool}}=5$; code time limit $\tau=60$s
\State $\mathcal{A}_0 \gets \varnothing$ \Comment{answers from prior rounds}
\For{$r=1,2,\dots,r_{\max}$}
  \Statex \textbf{// Round-$r$ message passing: each agent reads $q$ + prior answers}
  \State \textbf{parallel for} each $s\in\mathcal{S}$ \label{line:parallel}
    \State $p_r^s \gets \textsc{BuildPrompt}(q,\mathcal{A}_{r-1})$ \Comment{concatenate $q$ with all answers from prior round}
    \State $y_r^s \gets \textsc{AgentCall}(s,p_r^s,R_{\text{tool}},\tau)$ \Comment{tool-augmented reasoning (Alg.~\ref{alg:agentcall})}
  \State $\mathcal{A}_r \gets \{y_r^s : s\in\mathcal{S}\}$
  \If{$r \ge r_{\min}$ \textbf{and} \textsc{LLMTerminate}$(q,\mathcal{A}_{r})=\text{STOP}$} \label{line:terminate}
     \State \textbf{break}
  \EndIf
\EndFor
\State $a^\star \gets \textsc{MajorityVote}(\mathcal{A}_r)$
\State \Return $a^\star$
\end{algorithmic}
\end{algorithm}

\begin{algorithm}[!ht]
\caption{\textsc{AgentCall}$(s,p,R_{\text{tool}},\tau)$: tool-augmented reasoning for agent $s$ (returns answer only)}
\label{alg:agentcall}
\begin{algorithmic}[1]
\State $h \gets p$; $b \gets 0$ \Comment{$h$ is the running context; $b$ counts tool interactions}
\While{$b < R_{\text{tool}}$}
   \State $o \gets \textsc{LLMGenerate}(s,h)$ \Comment{run agent $s$ with its strategy/prompt}
   \If{$o$ contains a final answer $y$}
      \State \Return $o$
   \ElsIf{$o$ proposes code and $s$ allows \texttt{Code Interpreter}}
      \State $r \gets \textsc{ExecuteCode}(o,\tau)$ \Comment{hard limit $\tau=60$s; capture stdout/plots/files}
      \If{\textsc{RuntimeError}$(r)$}
         \State $h \gets h \parallel \text{``Runtime error:'' } r$; $b \gets b+1$; \textbf{continue}
      \Else
         \State $h \gets h \parallel \text{``Code result:'' } r$; $b \gets b+1$; \textbf{continue}
      \EndIf
   \ElsIf{$o$ issues a search query and $s$ allows \texttt{Search}}
      \State $E \gets \textsc{WebSearch}(o)$ \Comment{supports \texttt{gs}/\texttt{llm}/\texttt{com} variants}
      \State $h \gets h \parallel \text{``Retrieved evidence:'' } E$; $b \gets b+1$; \textbf{continue}
   \Else
      \State $h \gets h \parallel \text{``Continue reasoning with current context.''}$ \Comment{encourage self-reflection}
   \EndIf
\EndWhile
\State $o \gets \textsc{LLMGenerate}(s,h)$ \Comment{budget exhausted; force a decision}
\State \Return $o$
\end{algorithmic}
\end{algorithm}

\newpage
\section{Prompts of TUMIX}
\label{appendix section: Prompts of TUMIX}

\begin{table}[h!]
  \caption{Prompt for answer refinement based on all agent answers in the previous round.}
  \label{tab:final-answer-prompt}
  \centering
  \small 
  \renewcommand{\arraystretch}{1.2} 
  
  \begin{tabularx}{\linewidth}{X} 
    \toprule
    \textbf{Task}: Decide the final answer based on the following answers from other agents.
    \vspace{0.5em} 

    \textbf{Question}: \\
    \textcolor{red}{\texttt{\{question\}}}
    \vspace{0.5em}

    \textbf{Candidate answers from several methods}: \\
    \textcolor{red}{\texttt{\{joined\_answers\}}}
    \vspace{0.5em}
    
    Based on the candidates above, analyze the question step by step and try to list all the careful points. In the end of your response, directly output the answer to the question with the format \texttt{<<<answer content>>>}. \\
    \bottomrule
  \end{tabularx}
\end{table}

\begin{table}[h!]
  \caption{Prompt for LLM-as-Judge of refinement termination.}
  \label{tab:consensus-assessment-prompt}
  \centering
  \small 
  \renewcommand{\arraystretch}{1.25} 
  
  \begin{tabularx}{\linewidth}{X} 
    \toprule
    \textbf{Task}: Carefully assess whether the answers below (enclosed by \texttt{<<< >>>}) show clear and strong consensus, or if another round of reasoning is needed to improve alignment.
    \vspace{0.5em}

    {\color{AnswerRed}\textbf{IMPORTANT}: If there are any differences in reasoning, phrasing, emphasis, conclusions, or interpretation of key details, you should conservatively decide to continue refinement.}
    \vspace{0.5em}

    The current round number is \texttt{\{round\_num\}}. Note: {\color{AnswerRed}\textbf{Finalizing before round 3 is uncommon and discouraged unless answers are fully aligned in both logic and language.}}
    \vspace{1em}

    \textbf{Question}: \\
    \texttt{\{question\}}
    \vspace{1em}

    \textbf{Candidate answers from different methods}: \\
    \texttt{\{joined\_answers\}}
    \vspace{0.5em}

    \textbf{Instructions}:
    \begin{enumerate}[leftmargin=*, topsep=0.5em, itemsep=0.2em]
        \item Identify any differences in wording, structure, or logic.
        \item Be especially cautious about subtle variations in conclusion or emphasis.
        \item Err on the side of caution: if there's any ambiguity or divergence, recommend another round.
    \end{enumerate}
    \vspace{0.5em}
    
    Output your reasoning first, then conclude clearly with {\color{SearchCyan}\texttt{<<<YES>>>}} if the answers are highly consistent and finalization is safe, or {\color{AnswerRed}\texttt{<<<NO>>>}} if further refinement is needed. \\
    \bottomrule
  \end{tabularx}
\end{table}

\begin{table}[h!]
  \caption{Head prompt for \texttt{CoT Agent}.}
  \label{tab:text-output-prompt}
  \centering
  \small 
  \renewcommand{\arraystretch}{1.2} 
  
  \begin{tabularx}{\linewidth}{X} 
    \toprule
    \begin{itemize}[leftmargin=*, topsep=0.5em, itemsep=0.2em]
        \item Analyze the question step by step and try to list all the careful points.
        \item Then try to acquire the final answer with step by step analysis.
        \item In the end of your response, directly output the answer to the question.
    \end{itemize}
    \vspace{0.5em}

    \textbf{Do not output the code for execution.} \\
    \bottomrule
  \end{tabularx}
\end{table}

\begin{table}[h!]
  \caption{Head prompt for \texttt{CoT-Code Agent}.}
  \label{tab:COT-code-output-prompt}
  \centering
  \small 
  \renewcommand{\arraystretch}{1.2} 
  
  \begin{tabularx}{\linewidth}{X} 
    \toprule
    You are a helpful AI assistant. Solve tasks using your coding skills.
    \vspace{0.5em}

    In the following cases, suggest python code (in a python coding block) for the user to execute.
    
    \begin{itemize}[leftmargin=*, topsep=0.5em, itemsep=0.2em]
        \item Don't include multiple code blocks in one response, \textbf{only include one} in the response.
        \item Do not ask users to copy and paste the result. Instead, use the \texttt{'print'} function for the output when relevant.
    \end{itemize}
    \vspace{-0.5em} 

    Think the task step by step if you need to. If a plan is not provided, explain your plan first. You can first output your thinking steps with texts and then the final python code.
    \vspace{0.5em}

    \textbf{Remember in the final code you still need to output each number or choice in the final print!}
    \vspace{0.5em}
        
    Start the python block with {\color{SearchCyan}\texttt{\`{}\`{}\`{}python}} \\
    \bottomrule
  \end{tabularx}
\end{table}

\begin{table}[h!]
  \caption{Head prompt for \texttt{Code Agent}.}
  \label{tab:R1-Code-Interpreter-prompt}
  \centering
  \small
  \renewcommand{\arraystretch}{1.15}
  \begin{tabularx}{\linewidth}{X}
    \toprule
    The User asks a question, and you solve it. You first generate the reasoning and thinking process and then provide the User with the final answer. During the thinking process, **you can generate python code** for efficient searching, optimization, and computing with the format of starting the python block with {\color{SearchCyan}\texttt{\`{}\`{}\`{}python}}. {\color{InfoBrown}\texttt{**A code query must involve only a single script that uses `print' function for the output.**.}} Once the code script is complete, stop the generation. Then, the code interpreter platform will execute the code and return the execution output and error. Once you feel you are ready for the final answer, directly return the answer with the format {\color{AnswerRed}\texttt{$<<<$answer content$>>>$}} at the end of your response. Otherwise, you can continue your reasoning process and possibly generate more code query to solve the problem.\\
    \bottomrule
  \end{tabularx}
\end{table}

\begin{table}[h!]
  \caption{Head prompt for \texttt{Dual-Tool Agent}.}
  \label{tab:R1-Code-Interpreter-Search-prompt}
  \centering
  \small 
  \renewcommand{\arraystretch}{1.25} 
  
  \begin{tabularx}{\linewidth}{X} 
    \toprule
    The User asks a question, and you solve it. You first generate the reasoning and thinking process and then provide the User with the final answer.
    \vspace{1em} 

    During the thinking process, {\color{SearchCyan}you can generate python code} for efficient searching, optimization, and computing with the format of starting the python block with {\color{SearchCyan}\texttt{\`{}\`{}\`{}python}}. {\color{InfoBrown}\texttt{**A code query must involve only a single script that uses `print' function for the output.**.}}. Once the code script is complete, stop the generation. Then, the code interpreter platform will execute the code and return the execution output and error.
    \vspace{1em}

    If you lack the related knowledge, you can use the Google Search Tool to search the web and get the information. You can call a search query with the format of {\color{SearchCyan}\texttt{<search>your search query</search>}}, e.g., \texttt{<search>Who is the current president of US?</search>}. The searched results will be returned between \texttt{<information>} and \texttt{</information>}. Once the search query is complete, stop the generation. Then, the search platform will return the searched results.
    \vspace{1em}

    If you need to search the web, \textbf{do not generate code in the same response. Vice versa}. You can also solve the question without code and searching, just by your textual reasoning.
    \vspace{1em}

    Once you feel you are ready for the final answer, directly return the answer with the format {\color{AnswerRed}\texttt{<<<answer content>>>}} at the end of your response. Otherwise, you can continue your reasoning process and possibly generate more code or search queries to solve the problem. \\
    \bottomrule
  \end{tabularx}
\end{table}

\begin{table}[htbp]
  \caption{Head prompt for \texttt{Guided Agent}.}
  \label{tab:code-search-text-choice-prompt}
  \centering
  \small 
  \renewcommand{\arraystretch}{1.25} 
  
  \begin{tabularx}{\linewidth}{X} 
    \toprule
    You are guiding another TaskLLM to solve a task. You will be presented with a task that can be solved using textual reasoning, coding, and web searching. Sometimes the TaskLLM may need extra help to solve the task, such as generating code or searching the web. Then must follow the rules below for both query and return answer:
    \vspace{1em} 

    During the thinking process, {\color{SearchCyan}you can generate python code} for efficient searching, optimization, and computing with the format of starting the python block with {\color{SearchCyan}\texttt{\`{}\`{}\`{}python}}. {\color{InfoBrown}\texttt{A code query must involve only a single script that uses 'print' function for the output.}}. Once the code script is complete, stop the generation. Then, the code interpreter platform will execute the code and return the execution output and error.
    \vspace{1em}

    If you lack the related knowledge, you can use the Google Search Tool to search the web and get the information. You can call a search query with the format of {\color{SearchCyan}\texttt{<search>your search query</search>}}, e.g., \texttt{<search>Who is the current president of US?</search>}. The searched results will be returned between \texttt{<information>} and \texttt{</information>}. Once the search query is complete, stop the generation. Then, the search platform will return the searched results.
    \vspace{1em}

    If you need to search the web, \textbf{do not generate code in the same response. Vice versa}. You can also solve the question without code and searching, just by your textual reasoning.
    \vspace{1em}

    Once you feel you are ready for the final answer, directly return the answer with the format {\color{AnswerRed}\texttt{<<<answer content>>>}} at the end of your response. Otherwise, you can continue your reasoning process and possibly generate more code or search queries to solve the problem.
    \vspace{0.5em}

    \textbf{Your goal is to determine which method will be most effective for solving the task.} Then you generate the guidance prompt for the TaskLLM to follow in the next round. The final returned guidance prompt should be included between {\color{AnswerRed}\texttt{<<<}} and {\color{AnswerRed}\texttt{>>>}}, such as {\color{AnswerRed}\texttt{<<<You need to generate more complex code to solve...>>>}}.
    \vspace{1em}

    Now, here is the task: \\
    \bottomrule
  \end{tabularx}
\end{table}

\newpage
\section{Baseline methods}
\label{appendix sec: Baseline methods}

\begin{table}[htbp]
\centering
\caption{Baseline methods compared against \texttt{TUMIX}.}
\label{tab:baselines}
\footnotesize
\setlength{\tabcolsep}{3pt}
\renewcommand{\arraystretch}{1.05}
\begin{tabular}{@{}llp{0.64\linewidth}@{}}
\hline
\textbf{Method Handle} & \textbf{Type} & \textbf{Description} \\
\hline
\texttt{Majority-Vote} & Voting & A single agent runs multiple parallel inferences, with the final answer decided by majority voting, without sharing intermediate results. Uses \texttt{CS} agent. \\
\texttt{GSA} & Aggregation & Similar to \texttt{Majority-Vote}, but the same LLM generates a new response conditioned on multiple samples. Uses \texttt{CS} agent. \\
\texttt{Self-Reflection} & Iterative Refinement & A single agent iteratively refines its answer by reflecting on past responses (up to 10 accessible per round; varied to 8 or 15 in experiments, with no performance difference). Uses \texttt{CS} agent. \\
\texttt{SETS} & Multi-Trial Voting & The same LLM performs multiple self-reflection trials, and the final answer is chosen by majority vote. Uses \texttt{CS} agent. \\
\texttt{Self-MoA} & Best-Agent Selection & Selects the best-performing agent among 15 candidates for parallel sampling, answer sharing, and multi-round refinement. Adapted to select the best agent within the same LLM instead of the original setting, selecting the best LLM across different LLMs. \\
\texttt{Symbolic-MoE} & Expert Selection & Categorizes questions (e.g., algebra, probability, coding, biology), pre-tests the top 3 agents per category, and LLM judges the question category and assigns test questions to these top agents for sampling and aggregation. \\
\texttt{DEI} & Committee Heuristic & Selects the top 5 agents, generates multiple answers via repetitive sampling, and then uses a predefined agent committee with heuristics to select the best answer. \\
\texttt{SciMaster} & Critic and Refine & Samples the same pre-designed tool-use agent five times, then employs other agents to critique, refine, and aggregate the answers. Original prompts/agents retained. \\
\hline
\end{tabular}
\end{table}

\newpage
\section{Agent accuracy and coverage over multi-round answer refinement on HLE}
\label{appendix sec: agent accuracy evolution over rounds}

\begin{table}[ht]
\caption{Accuracy of each agent, average accuracy, and coverage across rounds for HLE. \texttt{Dual-Tool Agent}, \texttt{Guided Agent}, and \texttt{Guided Agent+} have three variants with different search strategies: Google Search API (\texttt{gs}), inherent LLM search function (\texttt{llm}), or their combination (\texttt{com}).}
\label{table: acc round evolution}
\centering
\renewcommand{\arraystretch}{1.3}
\setlength{\tabcolsep}{10pt}
\begin{tabular}{|>{\raggedright\arraybackslash}m{5cm}|c|c|c|c|c|c|}
\hline
\textbf{Humanity’s Last Exam (HLE)} & 
\textbf{RD 1} & 
\textbf{RD 2} & 
\textbf{RD 3} & 
\textbf{RD 4} &
\textbf{RD 5} &
\textbf{RD 6} \\
\hline
\textbf{Coverage} &
\textbf{51.92} & \textbf{44.20} & \textbf{43.48} & \textbf{37.04} & \textbf{34.85} & \textbf{33.87} \\
\hline
\textbf{Average} & \textbf{21.13} & \textbf{28.72} & \textbf{30.37} & \textbf{31.70} &
\textbf{32.16} &
\textbf{32.37} \\
\hline
\texttt{w/o TTS} & 20.32 & 28.08 & 30.88 & 31.60 & 32.18 & 32.36 \\
\hline
\texttt{CoT Agent (CoT)} & 20.84 & 28.16 & 30.40 & 31.12 & 32.30 & 32.48 \\
\hline
\texttt{CoT-Code Agent (CoT$_{\text{code}}$)} & 18.36 & 28.28 & 31.40 & 31.52 & 31.96 & 32.40 \\
\hline
\texttt{Search Agent (S)} & 21.72 & 29.04 & 28.84 & 32.08 & 32.08 & 32.20 \\
\hline
\texttt{Code Agent (C)} & 21.16 & 29.68 & 31.40 & 31.96 & 32.12 & 32.36 \\
\hline
\texttt{Code Agent+ (C$^{+}$)} & 22.96 & 29.00 & 31.44 & 31.88 & 32.20 & 32.40 \\
\hline
\texttt{Dual-Tool Agent (CS$_{\text{gs}}$)} & 22.96 & 28.60 & 30.84 & 31.72 & 32.24 & 32.36 \\
\hline
\texttt{Dual-Tool Agent (CS$_{\text{llm}}$)} & 21.36 & 28.36 & 31.28 & 31.20 & 32.48 & 32.32 \\
\hline
\texttt{Dual-Tool Agent (CS$_{\text{com}}$)} & 20.76 & 28.56 & 30.36 & 31.44 & 32.32 & 32.48 \\
\hline
\texttt{Guided Agent (CSG$_{\text{gs}}$)} & 22.04 & 28.72 & 29.96 & 32.24 & 32.00 & 31.96 \\
\hline
\texttt{Guided Agent (CSG$_{\text{llm}}$)} & 21.20 & 28.64 & 29.20 & 31.52 & 32.16 & 32.32 \\
\hline
\texttt{Guided Agent (CSG$_{\text{com}}$)} & 20.76 & 28.92 & 29.88 & 31.88 & 32.20 & 32.40 \\
\hline
\texttt{Guided Agent+ (CSG$^{+}$$_{\text{gs}}$)} & 20.56 & 29.32 & 30.36 & 31.92 & 31.84 & 32.52 \\
\hline
\texttt{Guided Agent+ (CSG$^{+}$$_{\text{llm}}$)} & 21.56 & 28.80 & 29.20 & 31.64 & 32.16 & 32.52 \\
\hline
\texttt{Guided Agent+ (CSG$^{+}$$_{\text{com}}$)} & 20.44 & 28.68 & 30.08 & 31.72 & 32.28 & 32.48 \\
\hline
\end{tabular}
\end{table}

\newpage
\section{Illustration and experimental results of \texttt{TUMIX} variants}
\label{appendix sec: Experimental results of TUMIX variants}

\begin{table}[!ht]
\centering
\caption{\texttt{TUMIX} framework and its variants, designed to ablate core components.}
\label{tab:tumix_variants}
\footnotesize
\setlength{\tabcolsep}{4pt} 
\renewcommand{\arraystretch}{1.2} 
\begin{tabular}{@{}llp{0.55\linewidth}@{}}
\toprule
\textbf{Method Handle} & \textbf{Component Ablated} & \textbf{Description} \\
\midrule
\texttt{TUMIX} & Main Method & (Default) Uses an LLM query to determine the optimal termination round (min. 2) and majority vote for final selection. \\
\addlinespace
\texttt{TUMIX-Rule} & Termination & Replaces the LLM-query termination with a rule: stops when the majority answer stabilizes across two consecutive rounds. \\
\addlinespace
\texttt{TUMIX-Fixed} & Termination & Replaces smart termination with a fixed 5-round limit, followed by majority voting for selection. \\
\texttt{TUMIX-FixedR} & Termination \& Selection & Uses a fixed 5-round limit, followed by random selection. \\
\addlinespace
\texttt{TUMIX-Evolve} & Agent Composition & Replaces the 15 human-designed agents with a static group of top-performing, LLM-generated agents for each refinement round. \\
\texttt{TUMIX-EvolveD} & Agent Composition & Extends the above by dynamically sampling a new agent group from the top-3 Evolved Agent groups for each refinement round. \\
\addlinespace
\texttt{TUMIX-Single} & Agent Diversity & Ablates diversity by replacing the 15 distinct agents with a single agent type from the $CS$ family. \\
\texttt{TUMIX-Three} & Agent Diversity & Reduces diversity by using only three agent types ($CS$, $C^+$, and $CSO$), each sampled 5 times per round. \\
\addlinespace
\texttt{TUMIX+} & Inference Scaling & Extends `TUMIX` with test-time scaling, running four inference passes per agent at different temperatures for the initial two rounds. \\
\bottomrule
\end{tabular}
\end{table}

\begin{table*}[!ht]
\caption{Experimental results of \texttt{TUMIX} variants. All the values are the average of three repetitive runs.}
\label{table: overall results TUMIX variants}
\vskip 0.15in
\begin{center}
\begin{small}
\begin{sc}
\begin{tabular}{lccccccccc}
\toprule
\multicolumn{1}{c}{Methods} & \multicolumn{9}{c}{\texttt{TUMIX} Variants}\\
\cmidrule(l){2-10}
\rotatebox{80}{Success rate \%} &
\rotatebox{80}{TUMIX} &
\rotatebox{80}{TUMIX-Single} &
\rotatebox{80}{TUMIX-Three} &
\rotatebox{80}{TUMIX-FixedR} &
\rotatebox{80}{TUMIX-Fixed} &
\rotatebox{80}{TUMIX-Rule} &
\rotatebox{80}{TUMIX-Evolve} &
\rotatebox{80}{TUMIX-EvolveD} &
\rotatebox{80}{TUMIX+}\\
\midrule
\multicolumn{1}{c}{} & \multicolumn{9}{c}{\textbf{Gemini-2.5-Pro}}\\
HLE & 32.3 & 29.0 & 30.2 & 32.4 & 32.4 & 32.4 & \textcolor{blue}{32.7} & 32.1 & \textcolor{magenta}{34.1} \\
GPQA & 87.9 & 86.1 & 86.6 & 86.8 & 86.7 & 87.7 & \textcolor{blue}{88.1} & 87.4 & \textcolor{magenta}{88.3} \\
AIME 24\&25 & \textcolor{blue}{96.7} & 95.0 & 95.3 & 95.6 & \textcolor{blue}{96.7} & \textcolor{blue}{96.7} & \textcolor{blue}{96.7} & 96.1 & \textcolor{magenta}{96.7} \\
\textbf{Ave. Norm.} & \textbf{72.3} & \textbf{70.0} & \textbf{70.7} & \textbf{71.6} & \textbf{71.9} & \textbf{72.3} & \textcolor{blue}{\textbf{72.5}} & \textbf{71.9} & \textcolor{magenta}{\textbf{73.0}} \\
\midrule
\multicolumn{1}{c}{} & \multicolumn{9}{c}{\textbf{Gemini-2.5-Flash}}\\
HLE & 21.2 & 18.2 & 18.6 & 20.9 & 20.8 & 21.3 & \textcolor{blue}{21.9} & 21.3 & \textcolor{magenta}{23.1} \\
GPQA & 77.3 & 65.8 & 67.1 & 76.8 & 77.1 & 77.4 & \textcolor{blue}{79.8} & 78.3 & \textcolor{magenta}{82.1} \\
AIME 24\&25 & 83.3 & 80.6 & 81.2 & 83.3 & 83.3 & 83.3 & \textcolor{blue}{86.7} & \textcolor{blue}{86.7} & \textcolor{magenta}{86.7} \\
\textbf{Ave. Norm.} & \textbf{60.6} & \textbf{54.9} & \textbf{55.6} & \textbf{60.3} & \textbf{60.4} & \textbf{60.7} & \textcolor{blue}{\textbf{62.8}} & \textbf{62.1} & \textcolor{magenta}{\textbf{64.0}} \\
\bottomrule
\end{tabular}
\end{sc}
\end{small}
\end{center}
\vskip -0.1in
\end{table*}

\newpage
\section{New agents in \texttt{TUMIX} completely designed by Gemini-2.5-Pro automatically}
\label{appendix section: new agents}
\begin{table}[h!]
\centering
\caption{Summary of 15 LLM-generated agents, categorized by their framework characteristics.}
\label{tab:baseline_agents}
\footnotesize
\setlength{\tabcolsep}{4pt}
\renewcommand{\arraystretch}{1.05}
\begin{tabular}{@{}llp{0.45\linewidth}@{}}
\hline
\textbf{Full Name} & \textbf{Short Name} & \textbf{Description} \\
\hline
\multicolumn{3}{@{}l}{\textbf{Iterative Agents (Multi-turn conversational frameworks)}} \\
\texttt{Plan-Verify-Refine} & \texttt{PVR} & Iteratively plans, executes one action (code or search), and refines based on checker feedback. \\
\texttt{SearchThenCode} & \texttt{S$\rightarrow$C} & Enforces a search-first, then code execution sequence in an iterative loop. \\
\texttt{CodeThenSearch} & \texttt{C$\rightarrow$S} & Enforces a code-first, then search execution sequence in an iterative loop. \\
\texttt{ConstraintPrune-Solver} & \texttt{CP$_{\text{solv}}$} & Iteratively prunes the solution space using constraints, guided by a checker and tools (code/search). \\
\texttt{MonteCarlo-Verify} & \texttt{MCV} & Uses Monte Carlo sampling via code to find a likely answer and then deterministically verifies it. \\
\texttt{Debate-CrossExam} & \texttt{DCE} & Simulates a Proposer/Skeptic debate to guide tool use, with a checker for cross-examination. \\
\texttt{MultiHop-Search-Aggregate} & \texttt{S$_{\text{m}}\rightarrow$C} & Enforces at least two sequential search actions before allowing any code execution. \\
\texttt{TDD-Code-Solver} & \texttt{TDD$_{\text{solv}}$} & A TDD agent that lists tests, writes code to pass them, and uses a checker for iterative refinement. \\
\hline
\multicolumn{3}{@{}l}{\textbf{Sequential Agents (Few-shot, non-conversational frameworks)}} \\
\texttt{SearchThenAnswer} & \texttt{S$\rightarrow$A} & A two-step agent that mandates a single web search before formulating the final answer. \\
\texttt{PlanThenCode} & \texttt{P$\rightarrow$C} & A two-step agent that first generates a plan, then a single code block to execute it. \\
\texttt{VerifierRefine} & \texttt{VR} & A three-step agent that generates a text answer, validates it with a checker, and then refines it. \\
\texttt{ToolSelector} & \texttt{TS} & Explicitly selects one tool (Search, Code, or Text) in the first step, then finalizes. \\
\texttt{HypothesisPruner-Code} & \texttt{HP$_{\text{code}}$} & Generates code to enumerate and prune solution hypotheses based on problem constraints. \\
\texttt{DualSearch-Consensus} & \texttt{S$^2_{\text{con}}$} & Issues two distinct search queries and then synthesizes the results into a consensus answer. \\
\texttt{TDD-CodeThenFix} & \texttt{TDD$_{\text{fix}}$} & A Test-Driven Development approach that writes tests and code, then generates a fix if tests fail. \\
\hline
\end{tabular}
\end{table}

\begin{table}[!ht]
\centering
\caption{Comparison of original agent group and top-3 agent group used in \texttt{TUMIX}, each with 15 agents, either pre-designed or LLM-generated.}
\label{tab:agent_group_original_top3}
\footnotesize
\setlength{\tabcolsep}{4pt}
\renewcommand{\arraystretch}{1.05}
\begin{tabular}{@{}lllp{0.25\linewidth}@{}}
\hline
\textbf{Original} & \textbf{Top-3-1} & \textbf{Top-3-2} & \textbf{Top-3-3}\\
\hline
\texttt{w/o TTS}        & \texttt{HypothesisPruner-Code} & \texttt{TDD-Code-Solver} & \texttt{w/o TTS} \\
\texttt{CoT Agent}      & \texttt{CoT Agent}  & \texttt{CoT Agent}  & \texttt{CoT Agent}  \\
\texttt{CoT-Code Agent} & \texttt{Plan-Verify-Refine}    & \texttt{CoT-Code Agent} & \texttt{CoT-Code Agent} \\
\texttt{Search Agent}   & \texttt{Search Agent}                      & \texttt{Search Agent}  & \texttt{SearchThenCode} \\
\texttt{Code Agent}     & \texttt{Code Agent}                      & \texttt{Code Agent}  & \texttt{TDD-Code-Solver} \\
\texttt{Code Agent+}    & \texttt{SearchThenCode}                      & \texttt{Code Agent+}  & \texttt{HypothesisPruner-Code}\\
\texttt{Dual-Tool Agent$_{\text{gs}}$}& \texttt{Dual-Tool Agent$_{\text{gs}}$}                      & \texttt{SearchThenCode}  & \texttt{DualSearch-Consensus} \\
\texttt{Dual-Tool Agent$_{\text{llm}}$}& \texttt{ConstraintPrune-Solver}                      & \texttt{Plan-Verify-Refine} & \texttt{MonteCarlo-Verify} \\
\texttt{Dual-Tool Agent$_{\text{com}}$}& \texttt{MonteCarlo-Verify}                      & \texttt{Dual-Tool Agent$_{\text{com}}$} & \texttt{ConstraintPrune-Solver}  \\
\texttt{Guided Agent$_{\text{gs}}$}   & \texttt{Guided Agent$_{\text{gs}}$}                     & \texttt{Guided Agent$_{\text{gs}}$}  & \texttt{Debate-CrossExam}\\
\texttt{Guided Agent$_{\text{llm}}$}   & \texttt{Guided Agent$_{\text{llm}}$}                     & \texttt{Guided Agent$_{\text{llm}}$}  & \texttt{Guided Agent$_{\text{llm}}$} \\
\texttt{Guided Agent$_{\text{com}}$}   & \texttt{Debate-CrossExam}                      & \texttt{Guided Agent$_{\text{com}}$}& \texttt{Guided Agent$_{\text{com}}$} \\
\texttt{Guided Agent+$_{\text{gs}}$}  & \texttt{Guided Agent+$_{\text{gs}}$}                     & \texttt{MonteCarlo-Verify} & \texttt{Guided Agent+$_{\text{gs}}$} \\
\texttt{Guided Agent+$_{\text{llm}}$}  & \texttt{SearchThenAnswer}                      & \texttt{Guided Agent+$_{\text{llm}}$} & \texttt{Plan-Verify-Refine}\\
\texttt{Guided Agent+$_{\text{com}}$}  & \texttt{DualSearch-Consensus}                      & \texttt{DualSearch-Consensus} & \texttt{Guided Agent+$_{\text{com}}$} \\
\hline
\end{tabular}
\vspace{-2mm}
\end{table}

\newpage
\section{Scaling behavior of Gemini-2.5-flash}
\label{appendix sec: Scaling behavior of Gemini-2.5-flash}
\begin{figure*}[ht]
  \centering
  \includegraphics[width=0.8\linewidth]{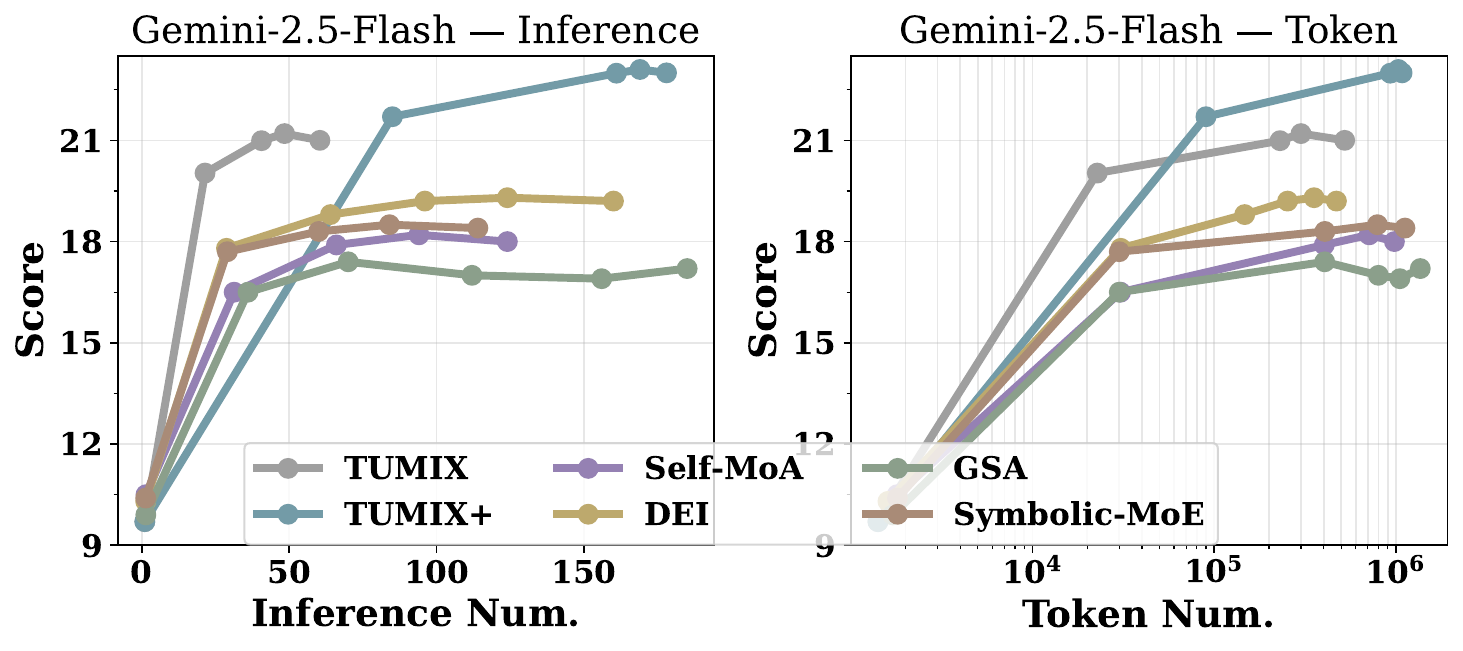}
   \caption{Scaling behavior of HLE scores in Gemini-2.5-flash relative to inference cost and total token count across different tool-augmented test-time scaling methods, where the token count includes both input and output tokens.}
   \label{fig:Scaling behavior of Gemini-2.5-flash}
\end{figure*}

\newpage
\section{LLM token confidence of generated responses}
\label{appendix sec: LLM token confidence of generated responses}
\begin{figure*}[ht]
  \centering
  \includegraphics[width=0.7\linewidth]{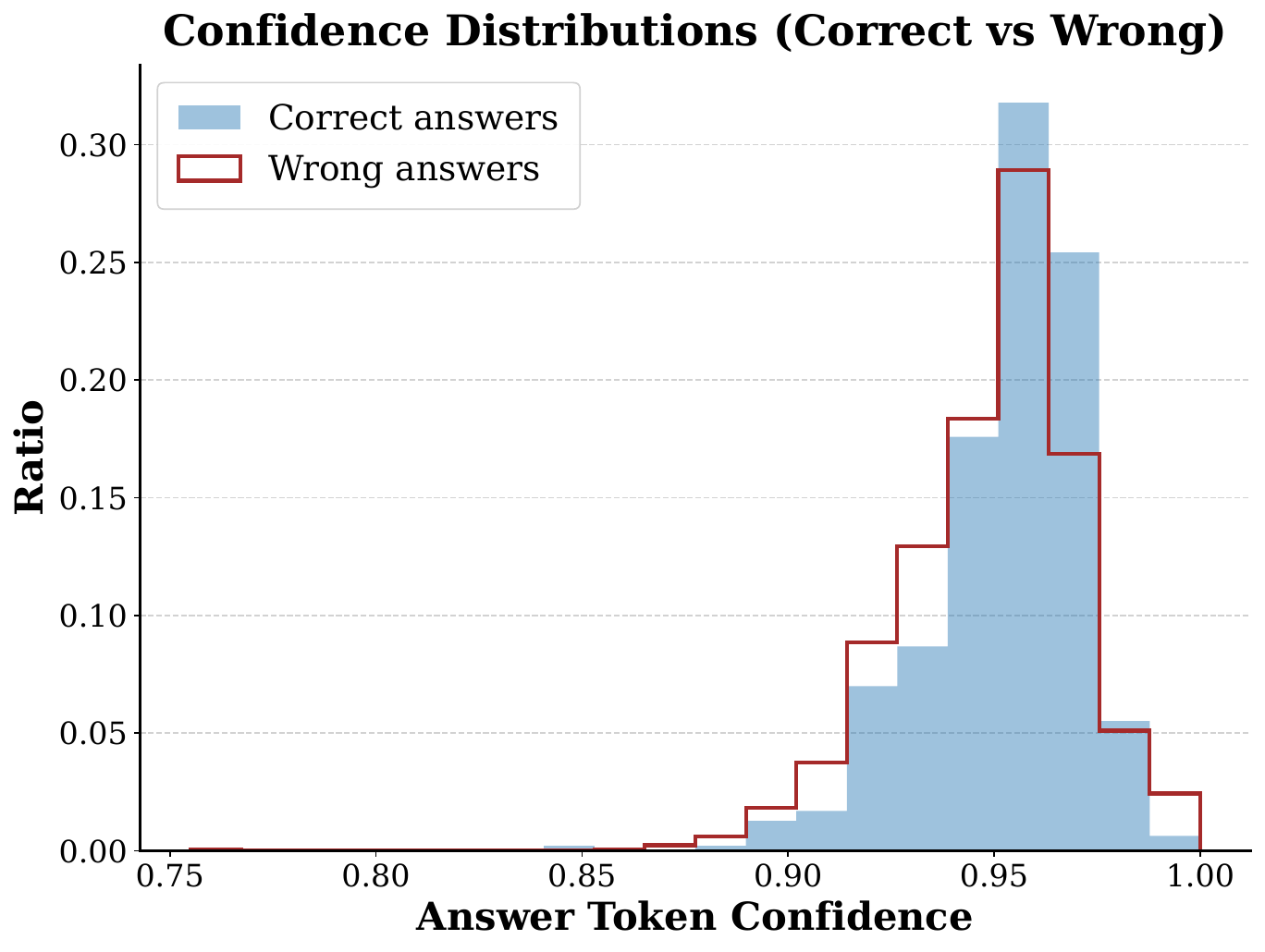}
   \caption{Distribution of LLM response confidence for correct and wrong answers. The response confidence is calculated based on the average token probability of the whole generated response. Here we use the responses of agent \texttt{CoT} as representative, as we find the distribution characteristics are very close among different agents.}
   \label{fig:logprob_compare}
\end{figure*}